\pdfoutput=1
\documentclass[twoside]{article}
\usepackage{times}
\usepackage[accepted]{aistats2016}
\usepackage{amssymb,amsmath,amsthm,stmaryrd}
\usepackage{natbib}

\usepackage{color}
\usepackage{hyperref}
\usepackage{algorithm}
\usepackage{algorithmic}
\usepackage{array}
\usepackage{graphicx} 
\usepackage{subfigure}
\usepackage{multicol}
\usepackage{tabulary}

\newtheorem{theorem}{Theorem}
\newtheorem{lemma}{Lemma}

\newtheorem{proposition}{Proposition}

\newtheorem{assumption}{Assumption}
\newtheorem{remark}{Remark}

\newcommand{\pr}[1]{\Pr\left[#1\right]}
\newcommand{\ort}[1]{\hat{#1}}
\newcommand{\td}[1]{\bar{#1}}
\newcommand{\Sk}{\mathcal{S}}
\newcommand{\Di}{\mathcal{D}}
\newcommand{\cO}{\mathcal{O}}
\newcommand{\R}{\mathbb{R}}

\newcommand{\E}[1]{\mathbb{E}\left[#1\right]}

\newcommand{\bx}{\mathbf{x}}
\newcommand{\bv}{\mathbf{v}}
\newcommand{\bu}{\mathbf{u}}

\definecolor{darkblue}{rgb}{0.0,0.0,0.7} 
\hypersetup{colorlinks,breaklinks, linkcolor=darkblue, urlcolor=darkblue,
            anchorcolor=darkblue, citecolor=darkblue}
\providecommand*\url[1]{\href{#1}{#1}} 
\renewcommand*\url[1]{\href{#1}{\texttt{#1}}} 

\begin{document}

\twocolumn[

\aistatstitle{Rivalry of Two Families of Algorithms for Memory-Restricted Streaming PCA}

\aistatsauthor{ Chun-Liang Li \And Hsuan-Ten Lin  \And Chi-Jen Lu }

\aistatsaddress{  \texttt{chunlial@cs.cmu.edu}\\ Carnegie Mellon University \And \texttt{htlin@csie.ntu.edu.tw} \\
National Taiwan University\And \texttt{cjlu@iis.sinica.edu.tw} \\ Academia Sinica } ]

\begin{abstract}

We study the problem of recovering the subspace spanned by the first $k$ principal components of $d$-dimensional data
under the streaming setting, with a memory bound of $\mathcal{O}(kd)$. Two families of algorithms are known for this
problem. The first family is based on the framework of stochastic gradient descent. Nevertheless, the convergence rate of the family can be seriously affected by the learning rate of the descent steps and deserves more serious study.
The second
family is based on the power method over blocks of data, but setting the block size for its existing algorithms is not
an easy task.  In this paper, we analyze the convergence rate of a representative algorithm with decayed learning
rate~\citep{oja} in the first family for the general $k>1$ case. Moreover, we propose a novel algorithm for the second
family that sets the block sizes automatically and dynamically with faster convergence rate.  We then conduct empirical
studies that fairly compare the two families on real-world data. The studies reveal the advantages and disadvantages of
these two families.

\end{abstract}

\section{Introduction}

For data points in
$\mathbb{R}^d$, the goal of principal component analysis (PCA) is to find the first $k \ll d$ eigenvectors  (principal components) that correspond to
the top-$k$ eigenvalues of the $d\!\times\!d$ covariance matrix.
For a batch of stored data points with a moderate $d$, efficient algorithms like the power method~\citep{m_comp} can be run on the empirical
covariance matrix to compute the solution.

In addition to the batch algorithms, the stream setting (streaming PCA) is attracting much research attention in recent
years~\citep{ACLS12, MCJ13, HP14}. Streaming PCA assumes that each data point $\mathbf{x} \in \mathbb{R}^{d}$ arrives
sequentially and it is not feasible to store all data points. If $d$ is moderate, the empirical covariance matrix can
again be computed and fed to an eigenproblem solver to compute the streaming PCA solution. When $d$ is huge, however, it
is not feasible to store the $\mathcal{O}(d^2)$ empirical covariance matrix. The situation arises in many modern
applications of PCA. Those applications call for memory-restricted streaming PCA, which will be the main focus of this
paper. We shall consider restricting to only $\mathcal{O}(kd)$ memory usage, which is of the same order as the minimum
amount needed for the PCA solution. In addition, we aim to develop streaming PCA algorithms that can keep improving the goodness
of the solution as more data points arrive. Such algorithms are free from a pre-specified goal of goodness and match
the practical needs better.

There are two measurements for the goodness of the solution.  One is the reconstruction error that
measures the expected squared error when projecting a data point to the solution, which is based on the fact that the
actual principal components should result in the lowest reconstruction error.  The other is the spectral error that
measures the difference between the subspace spanned by the solution and the subspace spanned by the actual principal
components, which will be formally defined in Section~\ref{sec:pre}. The spectral error enjoys a wide range of practical applications~\citep{SaRO15}. In addition, note that when the $k^{th}$ and $(k\!+\!1)^{th}$
eigenvalues are close, the solution that wrongly includes the $(k\!+\!1)^{th}$ engenvector instead of the $k^{th}$ one may still
reach a small reconstruction error, but the spectral error can be large.  That is, the spectral error is somewhat harder
to knock down and will be the main focus of this paper.

There are several existing streaming PCA algorithms, but not all of them focus on the spectral error \textit{and} meet
the memory restriction. For instance, \cite{KL15} proposed an algorithm which considers the spectral error, but
its space complexity is at least $\Omega(kd\log n)$, where $n$ is the number of data points received. Based on
\cite{WK08}, \cite{NKW13} proposed an algorithm along with regret guarantee on the reconstruction error, but
not the spectral error, and its space complexity grows in the order of $d^2$. \cite{ACS13} extended~\cite{ACLS12} to
derive convergence analysis for minimizing the reconstruction error along with a memory-efficient implementation,
but the space complexity is not precisely guaranteed to meet $\mathcal{O}(kd)$. That is, those works do not
match the focus of this paper.

There are two families of algorithms that tackle the spectral error while respecting the memory restriction, the family
of stochastic gradient descent (SGD) algorithms for PCA, and the family of block power methods.  The SGD family solves a
non-convex optimization problem that minimizes the reconstruction error, and applies SGD~\citep{oja} under the memory
restrictions to design streaming PCA algorithms. Interestingly, although the non-convex problem does not match standard
convergence assumptions of SGD~\citep{sgd}, minimizing the reconstruction error for the special case of $k = 1$ allows
\cite{fastpca} to derive spectral-error guarantees on the classic stochastic-gradient-descent PCA (SPCA)
algorithm~\citep{oja}.  
Recently, \cite{SaRO15} derive a spectral-error minimization algorithm for the general $k>1$
cases based on SGD along with strong theoretical guarantees. Nevertheless, different from~\cite{fastpca}, \cite{SaRO15} require a pre-specified error goal, which is taken to determine a \emph{fixed} learning rate of the descent step. The pre-specified goal makes the algorithm inflexible in taking more data points to further decrease the error. Furthermore, the fixed learning rate is inevitably conservative to keep the algorithm stable, but the conservative nature results in slow convergence in practice, as will be revealed from the experimental results in Section~\ref{sec:exp}.

The other family, namely the block power methods~\citep{MCJ13}, extends the batch power method~\citep{m_comp} for the memory-restricted streaming PCA
by defining blocks (periods) on the time line.  The key of the block power methods is to efficiently compute the product
of the estimated covariance matrices in different blocks. The product serves as an approximation to the power of the
empirical covariance matrix, which is a core element of the batch power method.  This family could also be viewed as
the mini-batch SGD algorithms but with different update rule from the SGD family.  The original block-power-method
PCA \citep[BPCA;][]{MCJ13} is proved to converge under some restricted distributions, 
which is later generalized by \cite{HP14} to a broader class of distributions. 
The convergence proof of BPCA in both works, however, depends on determining the block size from the total number of data points
or a pre-specified error goal, which again make the works inflexible for further decreasing the error with more data points. 

From the theoretical perspective, SPCA lacks convergence proof for the general $k > 1$ case without depending on the pre-specified error goal nor the fixed learning rate, and it is non-trivial to directly extend the fixed-learning-rate result of~\cite{SaRO15} to realize the proof;
from the algorithmic perspective, BPCA needs more algorithmic study on deciding the block size without depending on the pre-specified error goal;
from the practical perspective, it is not clear which family should be preferred in real-world applications.  This paper makes
contributions on all the three perspective. We first prove the convergence of SPCA for $k > 1$ with a decaying learning rate scheme
in Section~\ref{sec:sgd}. The convergence result turns out to be asymptotically similar to the result of \cite{SaRO15} while not relying on the fixed learning rate.
Then in Section~\ref{sec:bpm}, we propose a dynamic block power method (DBPCA) that automatically decides the block size
to not only allow easier algorithmic use but also guarantee better convergence rate. Finally, we conduct experiments on
real-world datasets and provide concrete recommendations in Section~\ref{sec:exp}.

\renewcommand{\arraystretch}{1.6}
\section{Preliminaries} \label{sec:pre}

Let us first introduce some notations which will be used later. First, let $x \le \cO(y)$ and $x \ge \Omega(y)$ denote
that for some universal constant $c$, independent of all our parameters, $x \le c y$ and $x \ge c y$, respectively, for
a large enough $y$. Next, let $\lceil x \rceil$ denote the smallest integer that is at least $x$. Finally, for a vector
$\bx$, we let $\|\bx\|$ denote its $\ell_2$-norm, and for a matrix $M$, we let $\|M\| = \max_{\bx}
\frac{\|M\bx\|}{\|\bx\|}$, which is the spectral norm.

In this paper, we study the streaming PCA problem, in which
with each input data point $\bx_n \in \R^d$ is received at step $n$ within a stream. Following previous works~\citep{MCJ13, fastpca}, we make the following assumption on the data distribution.
\begin{assumption} \label{as:A}
Assume that each $\bx_n$ is sampled independently from some distribution $\mathcal{X}$ with mean zero and covariance
matrix $A$, which has eigenvalues $\lambda_1\geq\lambda_2\geq \cdots \geq\lambda_d$, with $\lambda_i>\lambda_{i+1}$.
Moreover, assume that $\|\bx\| \leq 1$ for any $\bx$ in the support of $\mathcal{X}$,\footnote{We can relax this
  condition to that of having a small $\|\bx\|$ with a high probability as \cite{HP14} do, but we choose this stronger condition to simplify our presentation.} which implies that $\|A\| \le 1$ and $\|\bx_n \bx_n^\top \| \le 1$ for each $n$.
\end{assumption}
Our goal is to find a $d \times k$ matrix $Q_n$ at each step $n$, with its column-space quickly approaching that spanned by the first $k$ eigenvectors of $A$. For convenience, we let $\lambda=\lambda_k$ and $\ort{\lambda}=\lambda_{k+1}$, and moreover, let $U$ denote the $d\times k$ matrix with the first $k$ eigenvectors of $A$ as its columns.
One common way to measure the distance between such two spaces is
\begin{equation}\label{eq:Psi}
\Phi_n = \max_{\bv\in \R^k} \left(1-\frac{\|U^\top Q_n \bv\|^2}{\|\bv\|^2}\right),
\end{equation}
which can be used as an error measure for $Q_n$.
It is known that $\Phi_n = \sin\theta_k(U,Q_n)^2$, where $\theta_k(U,Q_n)$ is the $k$-th principle angle between these two spaces. For simplicity, we will denote $\sin\theta_k(U,Q_n)$ by $\sin(U,Q_n)$. Moreover, let $\cos(U,Q_n) = \sqrt{1-\sin(U,Q_n)^2}$ and $\tan(U,Q_n) = \sin(U,Q_n) / \cos(U,Q_n)$. It is also known that $\cos(U,Q_n)$ equals the smallest singular value of the matrix $U^\top Q_n$. More can be found in, e.g.,~\cite{m_comp}.

Our algorithms will generate an initial matrix $S_0 \in \R^{d\times k}$ by sampling each of its entries independently from the normal distribution $\mathcal{N}(0,1)$. Let $S_0 \sim \mathcal{N}(0,1)^{d\times k}$ denote this process, and we will rely on the following guarantee.

\begin{lemma} \citep{MCJ13} \label{lem:V_0}
Suppose we sample $S_0 \sim \mathcal{N}(0,1)^{d\times k}$ and let $S_0 = Q_0 R_0$ be its QR decomposition. Then for a large enough constant $\bar{c}$, there is a small enough constant $\delta_0$ such that
$\pr{\cos(U,Q_0) \le \sqrt{\bar{c}/(dk)}} \le \delta_0.$
\end{lemma}

\section{Stochastic Gradient Descent} \label{sec:sgd}

\begin{algorithm}[t]
  \caption{SPCA}
  \label{algo:oja}
  \fontsize{9pt}{9pt}
  \begin{algorithmic}[1]
    \STATE $S_0\sim \mathcal{N}(0, 1)^{d\times k}$
    \STATE $S_0 = Q_0 R_0$ (QR decomposition)
    \STATE $n \leftarrow 1$
    \WHILE{receiving data}
      \STATE $S_n \leftarrow Q_{n-1} + \gamma_n\bx_n\bx_n^\top Q_{n-1}$
      \STATE $S_n = Q_n R_n$ (QR decomposition)
      \STATE $n \leftarrow n+1$
    \ENDWHILE
  \end{algorithmic}
\end{algorithm}

In this section, we study the classic PCA algorithm framework of~\cite{oja} for the general rank-$k$ case,
which can be seen as performing stochastic gradient descent.
Our algorithm, called SPCA, is given in Algorithm~\ref{algo:oja}.
The key component is to determine the learning rate, which is related to the error analysis. 
We choose the step size at step $n$ as
  $$\gamma_n = \frac{c}{n}, \mbox{ with } c=\frac{c_0}{\lambda-\ort{\lambda}} \mbox{ for a constant } c_0 \ge 12.$$
The algorithm has a space complexity of $\cO(kd)$, by noting that the computation of $\bx_n\bx_n^\top Q_{n-1}$ can be done by first computing $\bx_n^\top Q_{n-1}$ and then multiplying the result by $\bx_n$. The sample complexity of our algorithm is guaranteed by the following, which we prove in Subsection~\ref{sec:itr}. Our analysis is inspired by and follows closely that of \cite{fastpca}
for the rank-one case, but there are several new hurdles which we need to overcome in the general rank-$k$ case.

\begin{theorem} \label{thm:itr_conv}
For any $\rho \in (0,1)$, there is some
$N \le (2^{1/(\lambda-\ort{\lambda})} kd)^{\cO(1)} + \cO\left(\frac{k\log (1/\rho)}{\rho(\lambda-\ort{\lambda})^3}\right),$ such that our algorithm with high probability can achieve $\Phi_n \le \rho$ for any $n \ge N$.
\end{theorem}

Let us remark that we did not attempt to optimize the first term in the bound above, as it is dominated by the second
term for a small enough $\rho$. Note that \cite{SaRO15} provided a better bound, which only has quadratic dependence of
the eigengap $\lambda-\ort{\lambda}$, for a similar algorithm called Alecton.  Alecton is restricted to taking a
\emph{fixed} learning rate that comes from a pre-specified error goal on a fixed amount of to-be-received data points.
The restriction makes Alecton less practical in the streaming setting, because one may not always be able to know the
amount of to-be-received data points in advance. If one receives fewer points than needed, Alecton cannot achieve the
error goal; if one receives more than needed, Alecton cannot fully exploit the additional points for a smaller error.
The decaying learning rate used by our proposed SPCA algorithm, on the other hand, does not suffer from such a
restriction.

\subsection{Proof of Theorem~\ref{thm:itr_conv}} \label{sec:itr}

The analysis of \cite{fastpca} works for the rank-one case by using a potential function
$\Psi_n = 1 - (U^\top Q_n)^2$, where $U$ and $Q_n$ are both vectors instead of matrices. To work in the general rank-$k$
case, we choose the function $\Phi_n$ defined in (\ref{eq:Psi}) as a generalization of their $\Psi_n$, and our goal
is to bound $\Phi_n$.

Following \cite{fastpca}, we divide the steps into epochs, with epoch $i$ ranging from step $n_{i-1}$ to step $n_i-1$, where we choose
    $n_0 = \hat{c}^c k^3 d^2 \log d$, for a large enough constant $\hat{c}$, and $n_i = \lceil e^{5/c_0} \rceil (n_{i-1}+1) -1$ for $i \ge 1$.
\begin{remark} \label{rm:c_1}
This gives us $(n_i+1) \ge e^{5/c_0} (n_{i-1}+1)$ and $n_i \le c_1 n_{i-1}$ for some constant $c_1$.
\end{remark}
As in \cite{fastpca}, we also use the convention of starting from step $n_0$.
For each epoch $i$, we would like to establish an upper bound $\rho_i$ on $\Phi_n$ for each step $n$ in that epoch.
To start with, we know the following from Lemma~\ref{lem:V_0}, using the fact that $\Phi_0 = \sin(U,Q_0)^2 = 1 - \cos(U,Q_0)^2$.
\begin{lemma} \label{lem:g0}
Let $\Gamma_0$ denote the event that $\Phi_0 \le \rho_0$, where $\rho_0 = 1 - \bar{c}/(kd)$ for the constant $\bar{c}$
in Lemma~\ref{lem:V_0}. Then we have $\pr{\neg \Gamma_0} \le \delta_0$.
\end{lemma}

Next, for each epoch $i\ge 1$, we consider the event
\[\Gamma_i: \sup_{n_{i-1} \le n < n_i} \Phi_n \le \rho_i,\]
for some $\rho_i$ to be specified later. Then our goal is to show that $\Pr[\neg \Gamma_{i+1} | \Gamma_i]$ is small, for
$i \ge 0$. This can be done for the rank-one case, but it relies crucially on the property that the potential function
$\Psi_n$ of \cite{fastpca} satisfies a nice recurrence relation.
Unfortunately, this does not appear so for our function $\Phi_n$, mainly because it takes an additional maximization over $v \in \R^k$. To overcome this problem, we take the following approach.

Consider an epoch $i$ and a step $n$ in the epoch. Let us define a new matrix
$Y_n = (I+\gamma_n\bx_n \bx_n^\top) Y_{n-1} = Q_n R_n R_{n-1} \cdots R_{n_{i-1}+1}$,
with $Y_{n_{i-1}} = Q_{n_{i-1}}$. Let $\Sk = \{\bv \in \R^k: \|\bv\|=1\}$.
Then for any $\bv \in \Sk$, define $$\Phi_n^{(\bv)} = 1 - \frac{\|U^\top Y_n \bv\|^2}{\|Y_n \bv\|^2},$$ and note that $\Phi_n =
\max_{\bv \in \Sk} \Phi_n^{(\bv)}$. Now for each such new function $\Phi_n^{(\bv)}$, with a fixed $\bv$, we can establish a similar recurrence relation as follows, but for our purpose later we show a better upper bound on $|Z_n|$ than that in
\cite{fastpca}. We give the proof in Appendix~\ref{app:rec}.

\begin{lemma} \label{lem:rec}
For any $n>n_0$ and any $\bv \in \Sk$, we have $\Phi_n^{(\bv)} \leq \Phi_{n-1}^{(\bv)} + \beta_n-Z_n$, where
\begin{enumerate}
  \item $\beta_n=5\gamma_n^2 + 2\gamma_n^3$
  \item  $|Z_n|\leq 2\gamma_n \sqrt{\Phi_{n-1}^{(\bv)}}$
  \item $\E{Z_n|\mathcal{F}_{n-1}}\geq 2\gamma_n(\lambda-\ort{\lambda}) \Phi_{n-1}^{(\bv)} (1-\Phi_{n-1}^{(\bv)})\geq
    0$.\footnote{As in \cite{fastpca}, $\mathcal{F}_{n-1}$ here denotes the $\sigma$-field of all outcomes up to and including step $n-1$.}
\end{enumerate}
\end{lemma}

With this lemma, the analysis of \cite{fastpca} can be used to show that $\mathbb{E}[\Phi_n^{(\bv)}]$ decreases as $n$ grows, but only for each individual $\bv$ separately. This alone is not sufficient to guarantee the event $\Gamma_{i+1}$
as it requires small $\Phi^{(\bv)}_n$ for all $\bv$'s simultaneously. To deal with this, a natural approach is to show
that each $\Phi^{(\bv)}_n$ is large with a small probability, and then apply a union bound, but an apparent difficulty
is that there are infinitely many $\bv$'s. We will overcome this difficulty by showing how it is possible to apply a
union bound only over a finite set of ``$\epsilon$-net" for these infinitely many $\bv$'s. Still, for this approach to
work, we need the probability of having a large $\Phi^{(\bv)}_n$ to be small enough, compared to the size of the $\epsilon$-net. However, the beginning steps of the first epoch seem to have us in trouble already as the probability of their $\Phi^{(\bv)}$ values exceeding $\Phi^{(\bv)}_{n_0}$ is not small. This seems to prevent us from having an error bound $\rho_1 < \rho_0$,
and without this to start, it is not clear if we could have smaller and smaller error bounds for later epochs. To handle this, we sacrifice the first epoch by using an error bound $\rho_1$ slightly larger than $\rho_0$, but still small enough. The hope is that once $\Gamma_1$ is established, we then have a period of small errors, and later epochs could then start to have decreasing $\rho_i$'s. More precisely, we have the following for the first epoch, which we prove in Appendix~\ref{app:g1}.

\begin{lemma} \label{lem:g1}
Let $\rho_1 = 1 - \bar{c}/(c_1^{6c} kd)$, for the constant $c_1$ given in Remark~\ref{rm:c_1}. Then $\pr{\neg \Gamma_1 \mid \Gamma_0} =0$.
\end{lemma}

It remains to set the error bounds for later epochs appropriately so that we can actually have small $\Pr[\neg \Gamma_{i+1} | \Gamma_i]$, for $i \ge 1$. We let the error bounds decrease in three phases as follows.
In the first phase, we let $\rho_i = 1-2(1-\rho_{i-1})$, so that $\eta_i = 1-\rho_i$ doubles each time. It ends at the first epoch $i$, denoted by $\pi_1$, such that $\rho_i < 3/4$. Note that $\pi_1\le \cO(\log d)$ and at this point, $\rho_{\pi_1}$ is still much larger than $1/n_{\pi_1}$.
Then in the second phase, we let $\rho_i = \rho_{i-1} / \lceil e^{5/c_0} \rceil^2$, which decreases in a faster rate
than $n_i$ increases. It ends at the first epoch $i$, denoted by $\pi_2$, such that $\rho_i \le c_2 (c^3 k \log
n_{i-1})/(n_{i-1}+1)$, for some constant $c_2$.\footnote{Determined later in the proof of Lemma~\ref{lem:ind} in
Appendix~\ref{sec:gamma} for the bound (\ref{eq:dev}) there to hold.} Note that $\pi_2\le \cO(\log d)$ and at this point, $\rho_{\pi_2}$ reaches about the order of $1/n_{\pi_2}$.
Finally in phase three, we let $\rho_i = c_2 (c^3 k \log n_{i-1})/(n_{i-1}+1)$, which decreases in about the rate as $n_i$ increases.

With these choices, the events $\Gamma_i$'s are now defined, and our key lemma is the following, which we prove in Appendix~\ref{sec:gamma}.
The proof handles the difficulties above by showing how a union bound can be applied only on a small ``$\epsilon$-net" of $\Sk$ along with proper choices of $\rho_i$ to guarantee that each $\Phi_n^{(\bv)}$ is large with a small enough probability.

\begin{lemma} \label{lem:ind}
For any $i\ge 1$, $\pr{\neg \Gamma_{i+1} \mid \Gamma_i} \le \frac{\delta_0}{2(i+1)^2}$.
\end{lemma}

From these lemmas, we can bound the failure probability of our algorithm as
\[
\begin{array}{lll}
  \pr{\exists i \ge 0: \neg \Gamma_j} &\le& \pr{\neg \Gamma_0} + \sum_{i\ge 0} \pr{\neg \Gamma_{i+1} \mid \Gamma_i} \\
  &\le& \delta_0 + \sum_{i\ge 0} \frac{\delta_0}{2(i+1)^2},
\end{array}
\]
which is at most $2 \delta_0$ using the fact that $\sum_{i \ge 1} 1/i^2 \le 2$.

To complete the proof, it remains to determine the number of samples needed by our algorithm to achieve an error bound $\rho$. This amounts to determine the number $n_i$ of an epoch $i$ with $\rho_i \le \rho$.
With $n_{\pi_2}\ge n_0$,
it is not hard to check that $\rho_{\pi_2} \le 1/ (2^c kd)^{\cO(1)}$ and $n_{\pi_2} \le (2^c kd)^{\cO(1)}$. Then if
$\rho \ge \rho_{\pi_2}$, we can certainly use $n_{\pi_2}$ as an upper bound. If $\rho \le \rho_{\pi_2}$, it is not hard
to check that with $n_i \le \cO(c^3 k(1/\rho)\log(1/\rho))$, we can have $\rho_i \le \rho$. As $c=c_0/(\lambda-\ort{\lambda})$, this proves Theorem~\ref{thm:itr_conv}.

\section{Block-Wise Power Method} \label{sec:bpm}

In this section, we turn to study a different approach based on block-wise power methods. Our algorithm is
modified from that of \cite{MCJ13} (referred as BPCA), which updates the estimate $Q_n$ with a more accurate estimate
of $A$ using a block of samples, instead of one single sample as in our first algorithm. Our algorithm differs from BPCA
by allowing different block sizes, instead of a fixed size. 
More precisely, we divide the
steps into blocks, with block $i$ consisting of steps from some interval $I_i$, and we use this block of $|I_i|$ samples
to update our estimate from $Q_{i-1}$ to $Q_i$. We will specify $|I_i|$ later in (\ref{eq:n_i}), which basically grows
exponentially after some initial blocks. We call our algorithm DBPCA, as described in Algorithm~\ref{algo:block}.

\begin{algorithm}[t]
  \caption{DBPCA}
  \label{algo:block}
  \fontsize{9pt}{9pt}
  \begin{algorithmic}[1]
    \STATE $S_0 \sim \mathcal{N}(0, 1)^{d\times k}$
    \STATE $S_0 = Q_0R_0$ (QR-decomposition)
    \STATE $i \leftarrow 1$
    \WHILE{receiving data}
      \STATE $S_i \leftarrow 0$
      \FOR{$n \in I_i$ }
       \STATE $S_i \leftarrow S_i+\frac{1}{|I_i|}\bx_n\bx_n^\top Q_{i-1}$
      \ENDFOR
      \STATE $S_i = Q_i R_i$ (QR-decomposition)
      \STATE $i \leftarrow i+1$
    \ENDWHILE
  \end{algorithmic}
\end{algorithm}

This algorithm, as our first algorithm SPCA, also has a space complexity of $\cO(kd)$. The sample complexity is guaranteed by
the following, which we will prove in Subsection~\ref{sec:pf}. To have a easier comparison with the results of
\cite{MCJ13} and \cite{HP14}, we use $\sqrt{\Phi_n} = \sin (U,Q_n)$ as the error measure in this section.

\begin{theorem} \label{thm:space}
Given any $\varepsilon \le 1/\sqrt{kd}$, our algorithm can achieve an error $\varepsilon$ with high probability after $L$ iterations with a total of $N$ samples, for some
$L \le \cO\left(\frac{\lambda}{\lambda-\ort{\lambda}}\log\frac{d}{\varepsilon}\right) \mbox{ and } N \le \cO\left(\frac{\lambda \log (dL)}{\varepsilon^2(\lambda-\ort{\lambda})^3}\right).$
\end{theorem}

Let us make some remarks about the theorem. First, the error $\rho$ in Theorem~\ref{thm:itr_conv} corresponds to the error $\varepsilon^2$ here,
and one can see that the bound in Theorem~\ref{thm:space} is better than those in
Theorem~\ref{thm:itr_conv} and \cite{MCJ13,HP14} in general. 
We summarize the sample complexity in terms of the error $\varepsilon$ in Table~\ref{tb:tb_comp_sgd}. Next, the condition $\varepsilon \le 1/\sqrt{kd}$ in the theorem is only used to simplify the error bound. One can check that our analysis also works for any $\varepsilon \le 1$, but the resulting bound for $N$ has the factor $\varepsilon^2$ replaced by $\min(1/(kd), \varepsilon^2)$. Finally, from Theorem~\ref{thm:space}, one can also express the error in terms of the number of samples $n$ as
$\varepsilon(n) \le \cO\left(\sqrt{\frac{\lambda}{n(\lambda-\ort{\lambda})^3} \log\left(\frac{\lambda d \log n}{\lambda-\ort{\lambda}}\right)}\right).$

\subsection{Proof of Theorem~\ref{thm:space}} \label{sec:pf}

Recall that after the $i$-th block, we have the estimate $Q_i$, and we would like it to be close to $U$, with a small
error $\sin(U,Q_i)$. To bound this error, we follow \cite{HP14} and work on bounding a surrogate error $\tan(U,Q_i)$, which suffices as $\sin(U,Q_i) \le \tan(U,Q_i)$.

To start with, we know from Lemma~\ref{lem:V_0} that for $\varepsilon_0 = \sqrt{\bar{c}/(kd)}$ with some constant $\bar{c}$,
$\Pr[\tan(U,Q_0) > \varepsilon_0] \le \delta_0,$
using the fact that $\tan(U,Q_0)^2 = 1/\cos(U,Q_0)^2 -1$.

Next, we would like to bound each $\tan(U,Q_i)$ in terms of the previous $\tan(U,Q_{i-1})$. For this, recall that with $F_i = \sum_{n \in I_i} \bx_n \bx_n^\top / |I_i|$, we have $Q_i R_i = F_i Q_{i-1}$, which can be rewritten as $A Q_{i-1} + (F_i - A)Q_{i-1}$. Using the notation $G_i = (F_i - A) Q_{i-1}$, we have
$Q_i R_i = A Q_{i-1} + G_i,$
where $G_i$ can be seen as the noise arising from estimating $A$ by $F_i$ using the $i$-th block of samples. Then, we
rely on the following lemma from \cite{HP14}, with the parameters:
\begin{equation}\label{eq:gamma}
  \td{\lambda} = \max(\ort{\lambda}, \lambda/4), \gamma = \left(\td{\lambda}/{\lambda}\right)^{1/4} \mbox{
  and }\triangle = (\lambda-\td{\lambda})/{4}.
\end{equation}

\begin{table}[t]
  \centering
    \footnotesize
    \begin{tabulary}{1.0\textwidth}{c|c|c}
    \hline
    Algorithm & Complexity & Restriction \\
    \hline \hline
    \multicolumn{3}{l}{SGD family}\\
    \hline
    \cite{fastpca} & $\cO( \frac{\log(1/\varepsilon)}{\varepsilon^2} )$ & only for $k=1$ \\
    \cite{SaRO15} & $\cO( \frac{\log(1/\varepsilon)}{\varepsilon^2} )$ & pre-specified $\varepsilon$ \\
    our proposed SPCA & $\cO( \frac{\log(1/\varepsilon)}{\varepsilon^2} )$ & none \\
    \hline\hline
    \multicolumn{3}{l}{block power method family}\\
    \hline
    \cite{HP14} & $\cO( \frac{\log(1/\varepsilon)}{\varepsilon^2} )$  & pre-specified $\varepsilon$ \\
    our proposed DBPCA & $\cO\left( \frac{\log(\log(1/\varepsilon))}{\varepsilon^2}\right)$ & none \\
    \hline
  \end{tabulary}
  \caption{sample complexity and restriction}
  \label{tb:tb_comp_sgd}
\end{table}

\begin{lemma} \citep{HP14} \label{lem:next}
Suppose $\|G\| \le \triangle \cdot \min(\cos(U,Q), \beta)$, for some $\beta >0$. Then $$\tan (U,AQ+G) \le \max(\beta, \max(\beta, \gamma) \tan(U,Q)).$$
\end{lemma}

From this, we can have the following lemma, proved in Appendix~\ref{app:err}, which provides an exponentially-decreasing upper bound on $\tan (U,Q_i)$, for the parameters:
$$\varepsilon_i = \varepsilon_0 \gamma^i \mbox{ and } \beta_i = \min\left(\gamma/\sqrt{1+\varepsilon_{i-1}^2}, \gamma
\varepsilon_{i-1}\right)$$
where $\varepsilon_0 = \sqrt{\bar{c}/(dk)}$ with the constant $\bar{c}$ in Lemma~\ref{lem:V_0}.

\begin{lemma} \label{lem:err}
Suppose $\tan (U,Q_{i-1}) \le \varepsilon_{i-1}$ and $\|G_i\| \le \triangle \beta_i$.
Then $\tan (U,Q_i) \le \varepsilon_i$.
\end{lemma}

The key which sets our approach apart from that of \cite{MCJ13,HP14} is the following observation. According to Lemma~\ref{lem:err}, for earlier iterations, one can in fact tolerate a larger $\|G_i\|$ and thus a larger empirical error for estimating $A$. This allows us to have smaller blocks at the beginning to save the number of samples, while still keeping the failure probability low. More precisely, we have the following lemma, proved in Appendix~\ref{app:dev}, with the parameters:
\begin{equation}\label{eq:n_i}
\delta_i = \delta_0/(2i^2) \mbox{ and } |I_i| = (c/(\triangle\beta_i)^2)\log(d/\delta_i),
\end{equation}
where $\delta_0$ is the error probability given in Lemma~\ref{lem:V_0} and $c$ is a large enough constant.

\begin{lemma} \label{lem:dev}
For any $i\ge 1$, given $|I_i|$ samples in iteration $i$, we have $\Pr[\|G_i\| > \triangle\beta_i] \le \delta_i$.
\end{lemma}

With this, we can bound the failure probability of our algorithm as
\[
\begin{array}{l}
  \pr{\exists i\ge 0: \tan(U,Q_i) >\varepsilon_i} \le \\
  \pr{\tan(U, Q_0) >\varepsilon_0} + \sum_{i\ge 1} \pr{\|G_i\| > \triangle\beta_i}
\end{array}
\]
which by Lemma~\ref{lem:V_0} and Lemma~\ref{lem:dev} is at most
$\delta_0 + \sum_{i\ge 1} \delta_i = \delta_0 + \sum_{i\ge 1} \frac{\delta_0}{2i^2} \le 2\delta_0.$

To complete the proof of Theorem~\ref{thm:space}, it remains to bound the number of samples needed for achieving error
$\varepsilon$.
For this, we rely on the following lemma which we prove in Appendix~\ref{app:n_sample}.

\begin{lemma} \label{lem:n_sample}
For some $L \le \cO\left(\frac{\lambda}{\lambda-\td{\lambda}}\log\frac{d}{\varepsilon}\right)$, we have $\varepsilon_L \le \varepsilon$
and
$\sum_{i=1}^L |I_i| \le \cO\left(\frac{\lambda\log (dL)}{\varepsilon^2(\lambda-\td{\lambda})^3}\right).$
\end{lemma}

Finally, as $\td{\lambda} = \max(\ort{\lambda}, \lambda/4)$, we have $\lambda-\td{\lambda} \ge
\Omega(\lambda-\ort{\lambda})$, and putting this into the bound above yields the sample complexity bound stated in the theorem.

\renewcommand\arraystretch{1}
\section{Experiment} \label{sec:exp}

We conduct experiments on two large real-world datasets NYTimes and PubMed~\citep{data} as used by~\cite{MCJ13}. The dimension $d$ of the data points in the datasets are $102$ and~$141$ thousands, respectively, which match our memory-restricted setting. The features of both datasets are normalized into $[0,1]$.

Parameter tuning is generally difficult for streaming algorithms. Instead of tuning the parameters extensively and
reporting with the most optimistic (but perhaps unrealistic) parameter choice for each algorithm, we consider a thorough
range of parameters but report the results of four parameter choices per algorithm, which cover the best parameter
choice, to understand each algorithm more deeply.

We compare the proposed SPCA and DBPCA with Alecton~\citep[fixed-learning-rate;][]{SaRO15} and BPCA~\citep[fixed-block-size;][]{HP14}.
For Alecton, we report the results of the learning rate $\gamma \in \{10^{-1}, \cdots, 10^3\}$, with reasons to be explained in Subsection~\ref{sec:alec}.
For SPCA, we follow its existing work~\citep{fastpca} to fix $n_0 = 0$ while considering $c \in \{10^1,\cdots, 10^8\}$. 
Then we report the results of $c \in \{10^3, 10^4, 10^5, 10^6\}$.  For BPCA, we follow its
existing works~\citep{MCJ13, HP14} and let the block size be $\lfloor N/T\rfloor$, where $N$ is the size of the dataset
and $T$ is the number of blocks. Theoretical results of BPCA~\citep{HP14} suggest $T =
\cO\left(\frac{\lambda}{\lambda-\ort{\lambda}}\log d\right)$. 
Because $\lambda$ and $\ort{\lambda}$ are unknown in
practice, \cite{MCJ13, HP14} set $T=\lfloor L\log d \rfloor$ with $L=1$.
Instead, we extend the range to $L \in \{5^{-1},\cdots, 5^{6}\}$ and report the result of $\{5^0, 5^1, 5^2, 5^3\}$.

For the proposed DBPCA, we set the initial block size as $2k$ to avoid being rank-insufficient in the first block. Then, we consider the ratio $\gamma^2 \in \{0.6, 0.7, 0.8, 0.9\}$
for enlarging the block size.
\footnote{Note that (\ref{eq:gamma}) suggests $\gamma^2\geq 0.5$.}

\begin{figure}
  \centering
  \subfigure[Alecton, $k=4$]{\label{fig:ny_a_4}\includegraphics[width=0.23\textwidth]{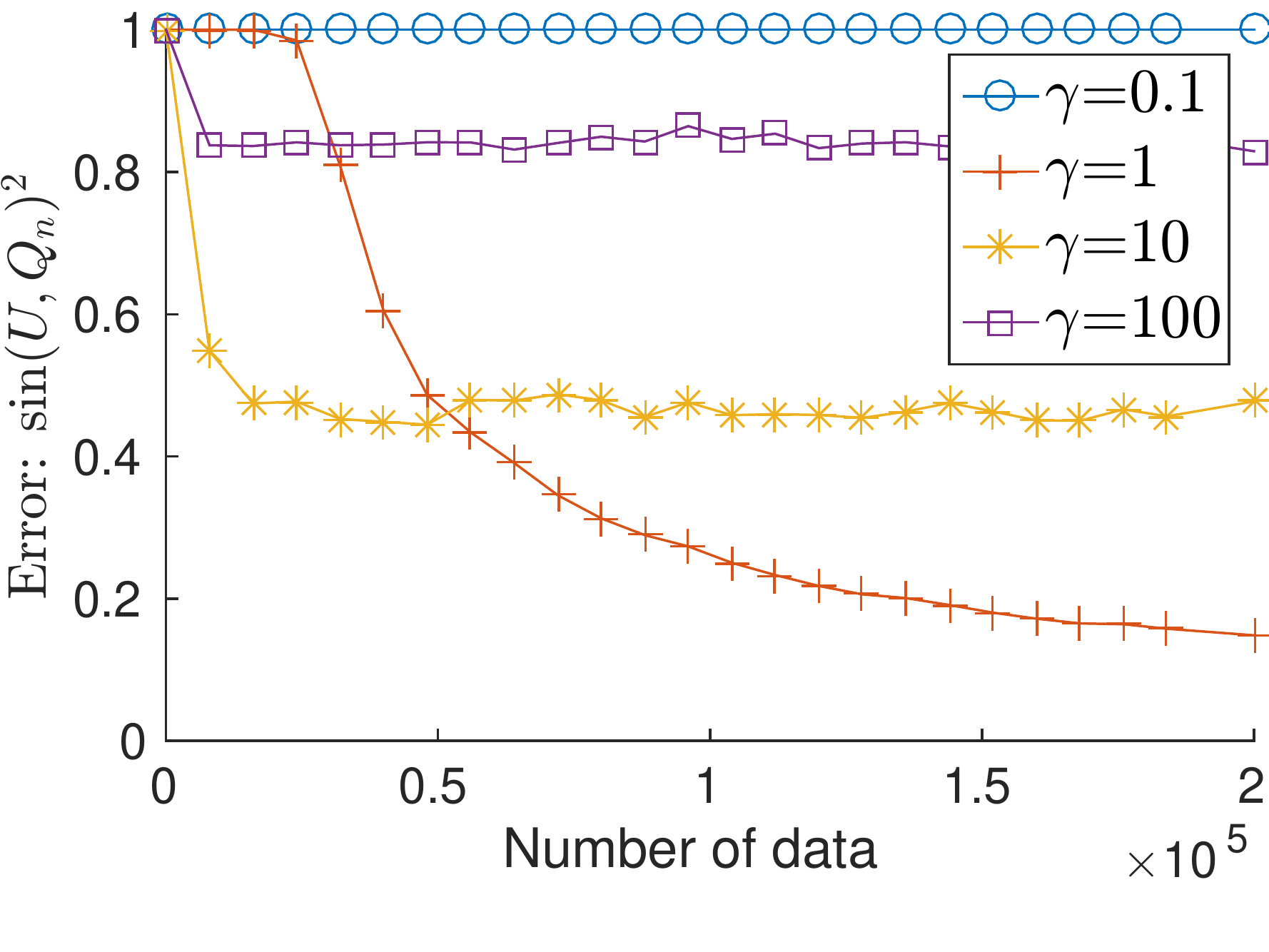} } 
  \subfigure[SPCA, $k=4$]{\label{fig:ny_s_4}\includegraphics[width=0.23\textwidth]{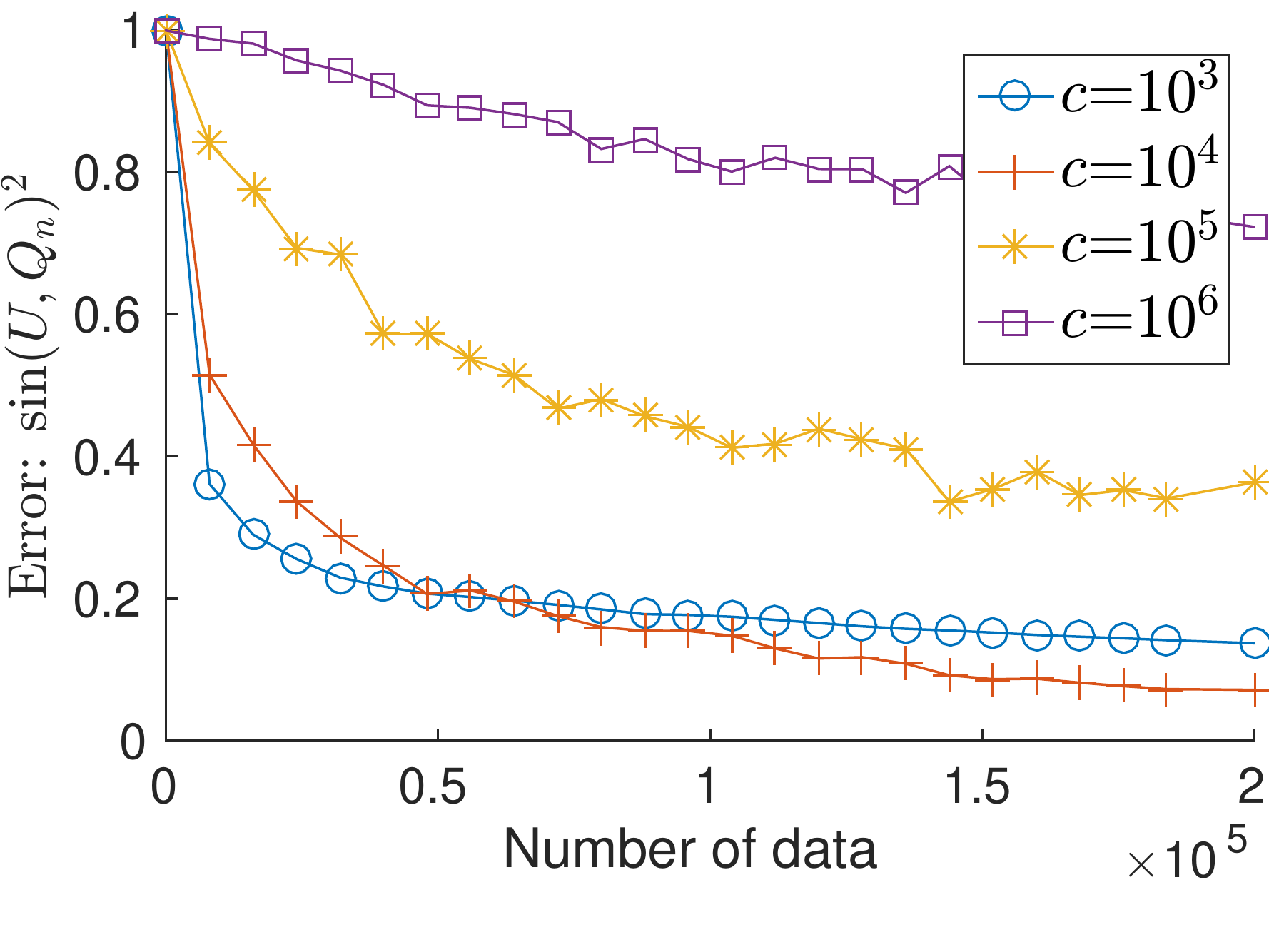} } 
  \subfigure[BPCA, $k=4$]{\label{fig:ny_b_4}\includegraphics[width=0.23\textwidth]{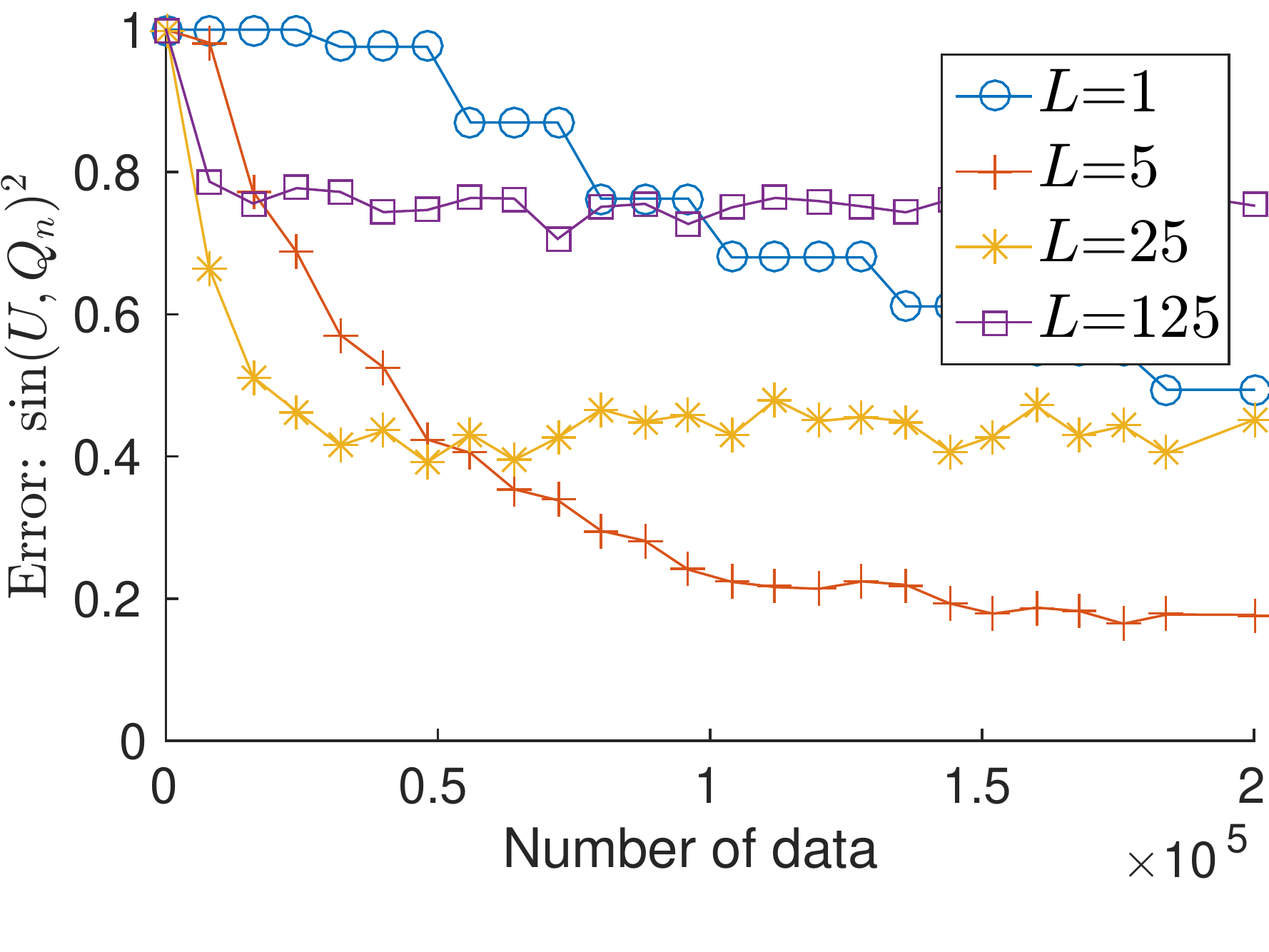} } 
  \subfigure[DBPCA, $k=4$]{\label{fig:ny_d_4}\includegraphics[width=0.23\textwidth]{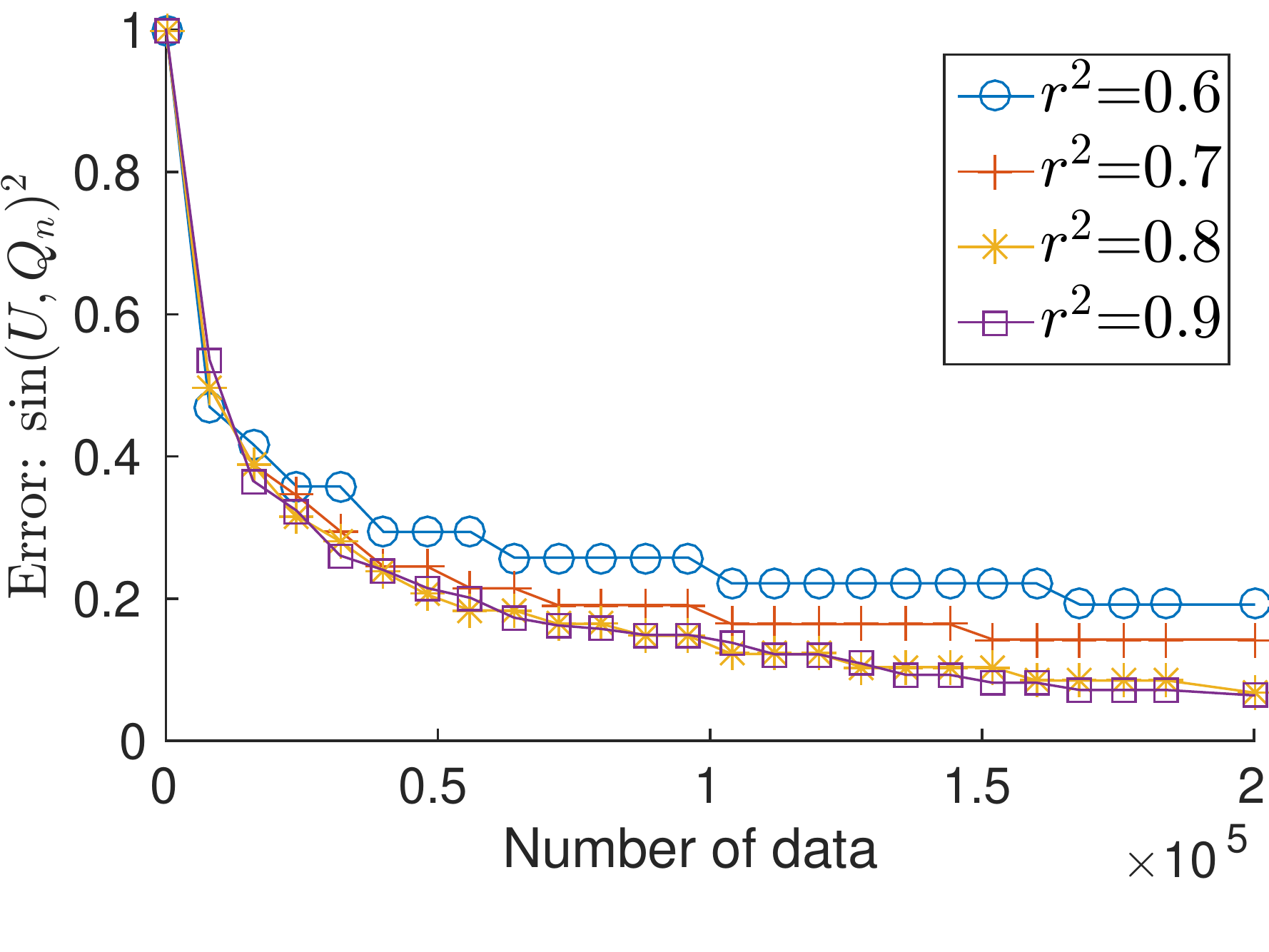} } 
  \caption{Performance of different algorithms on NYTimes when $k=4$}
  \label{fig:ny4}
\end{figure}

\begin{figure}[t]
  \centering
  \subfigure[Alecton, $k=10$]{\label{fig:ny_a_10}\includegraphics[width=0.23\textwidth]{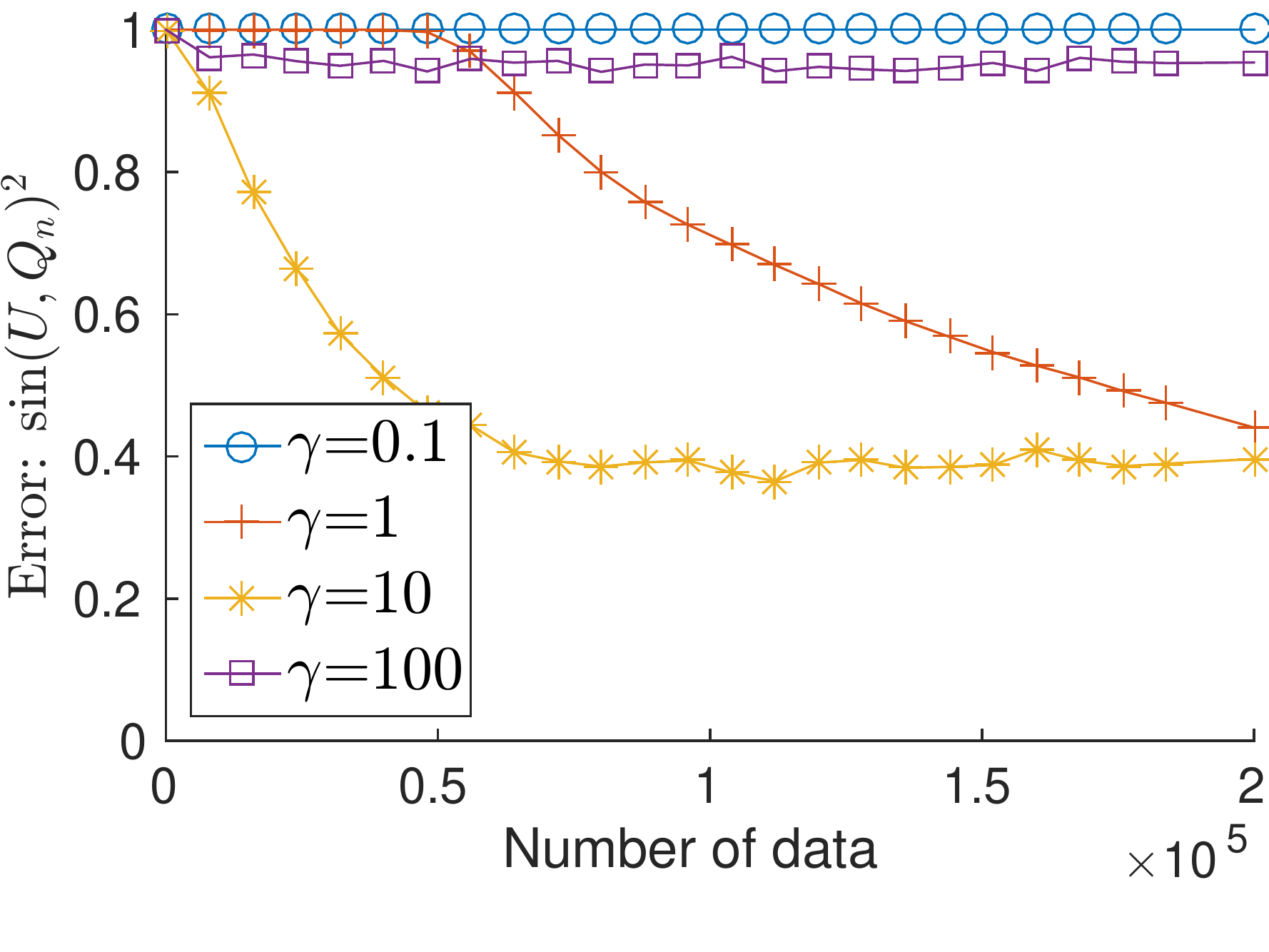} } 
  \subfigure[SPCA, $k=10$]{\label{fig:ny_s_10}\includegraphics[width=0.23\textwidth]{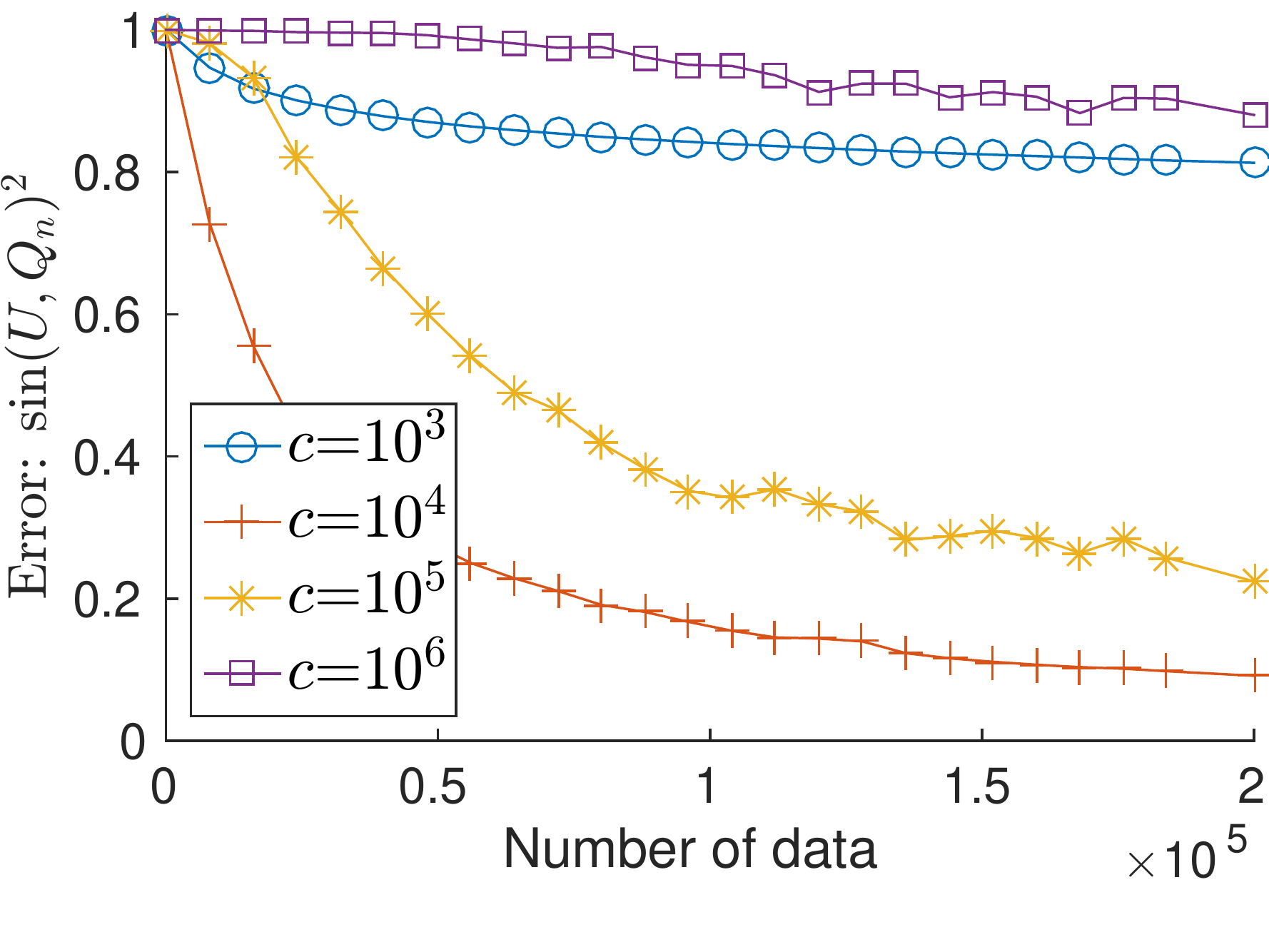} } 
  \subfigure[BPCA, $k=10$]{\label{fig:ny_b_10}\includegraphics[width=0.23\textwidth]{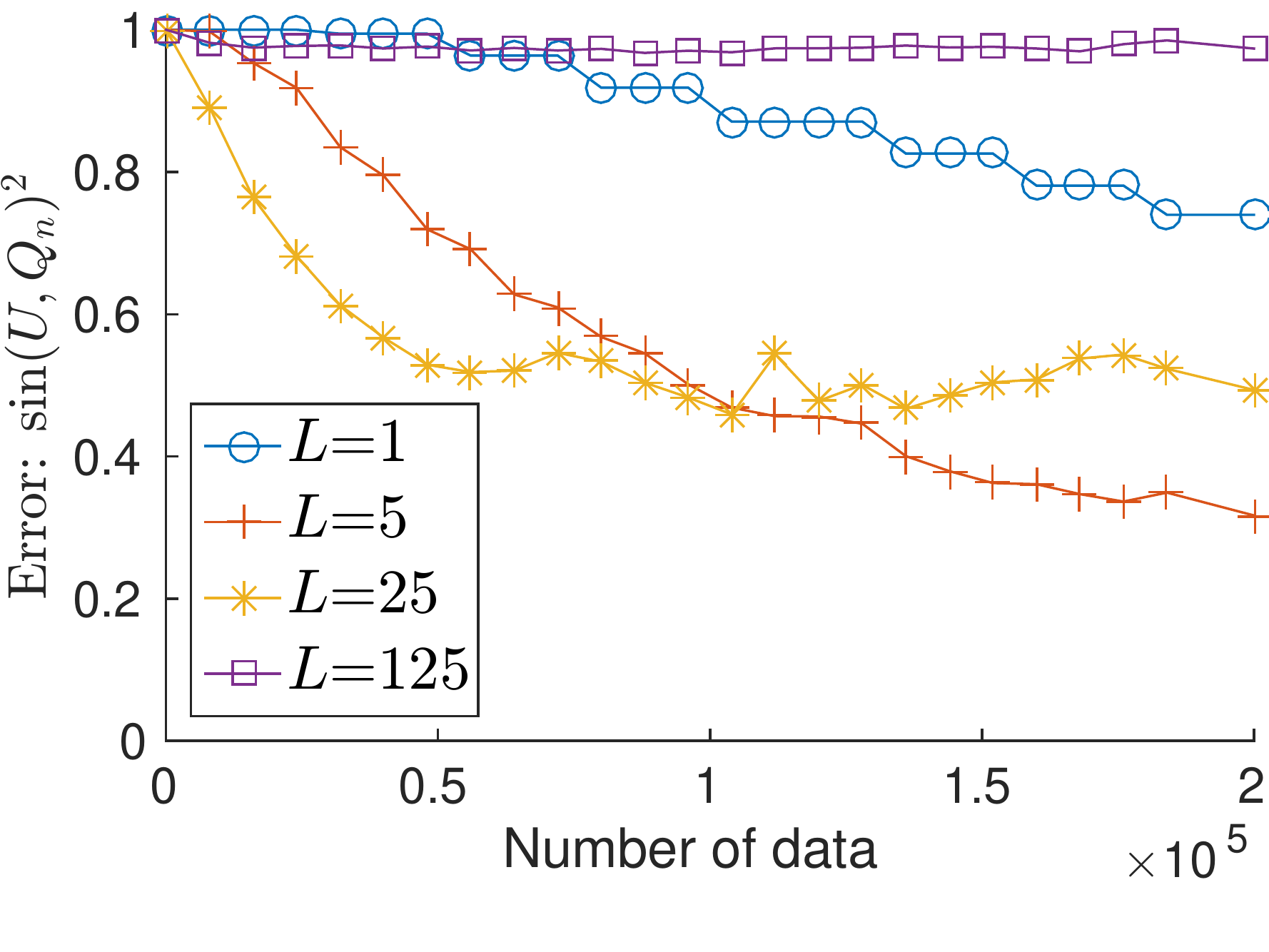} } 
  \subfigure[DBPCA, $k=10$]{\label{fig:ny_d_10}\includegraphics[width=0.23\textwidth]{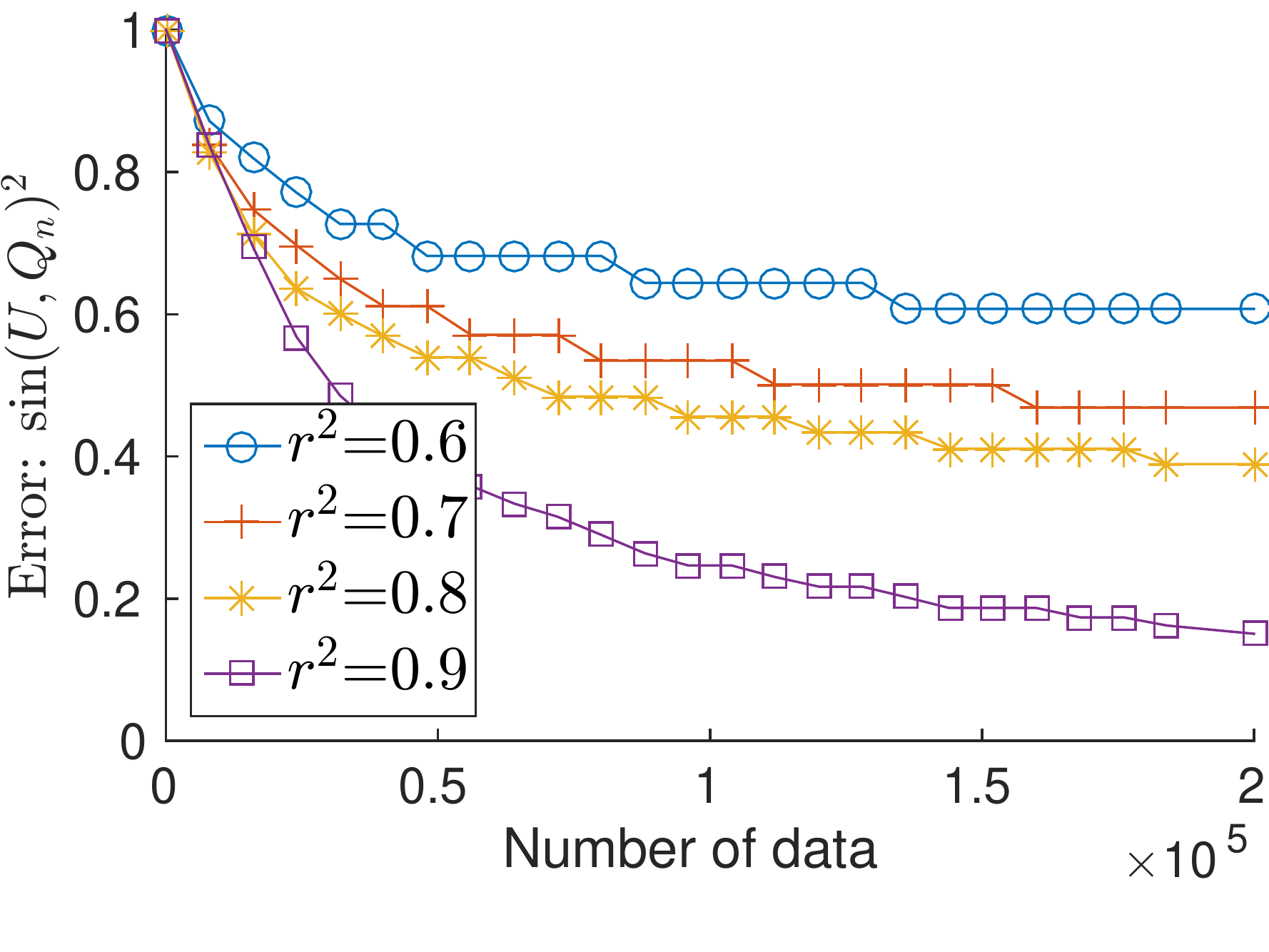} } 
  \caption{Performance of different algorithms on NYTimes when $k=10$}
  \label{fig:ny10}
\end{figure}

\begin{figure}[t]
  \centering
  \subfigure[Alecton, $k=4$]{\label{fig:pm_a_4}\includegraphics[width=0.23\textwidth]{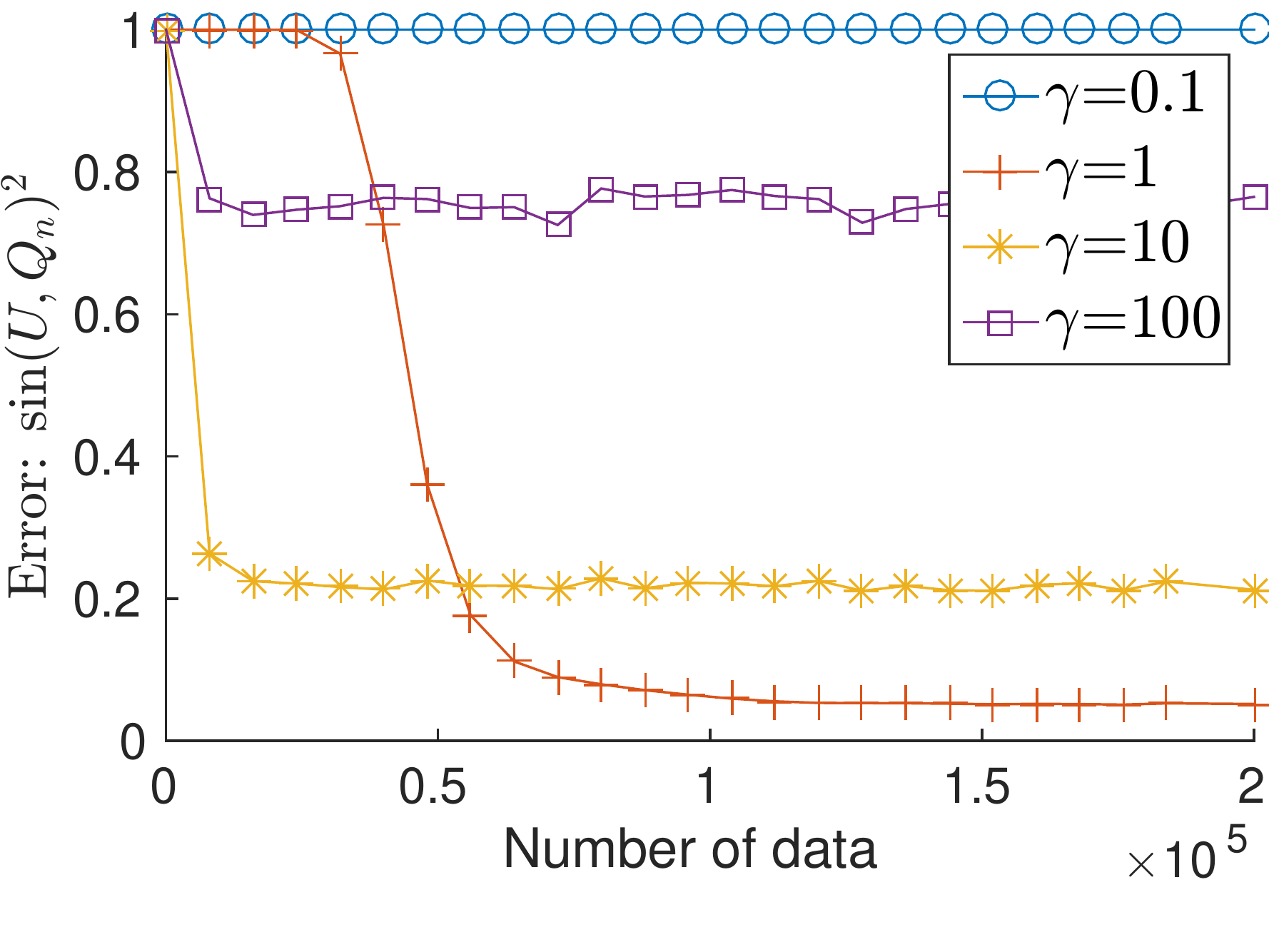} } 
  \subfigure[SPCA, $k=4$]{\label{fig:pm_s_4}\includegraphics[width=0.23\textwidth]{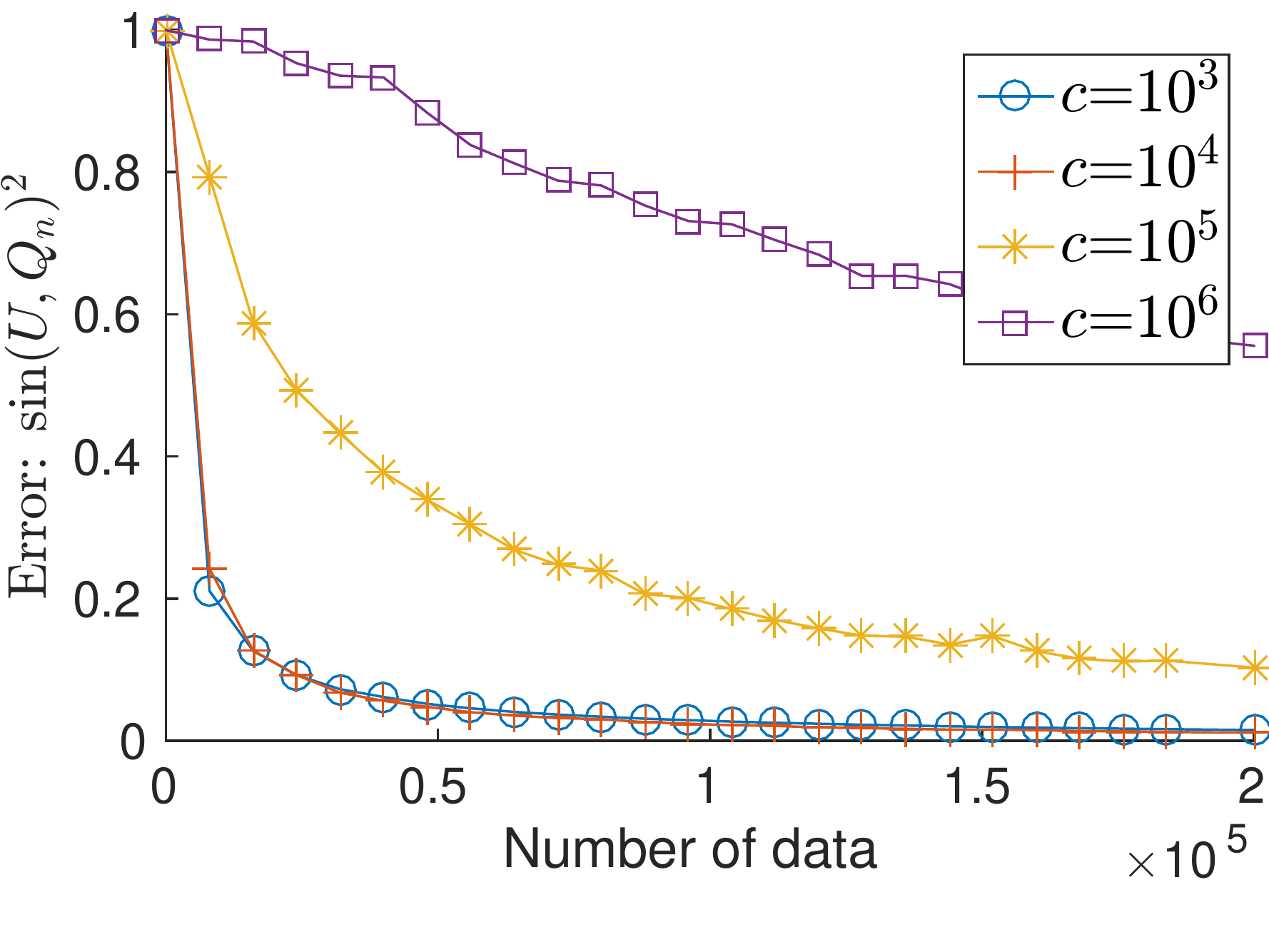} } 
  \subfigure[BPCA, $k=4$]{\label{fig:pm_b_4}\includegraphics[width=0.23\textwidth]{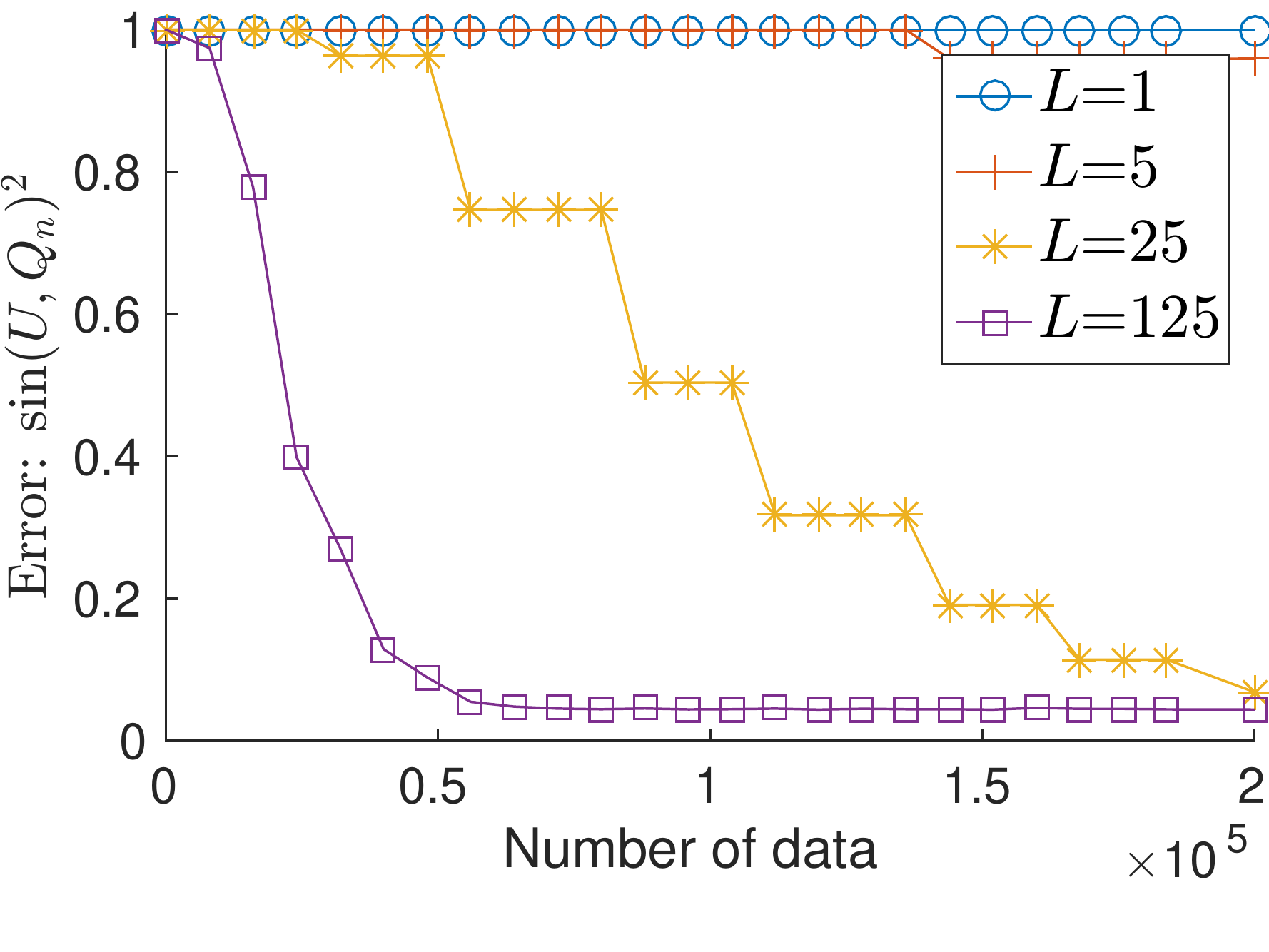} } 
  \subfigure[DBPCA, $k=4$]{\label{fig:pm_d_4}\includegraphics[width=0.23\textwidth]{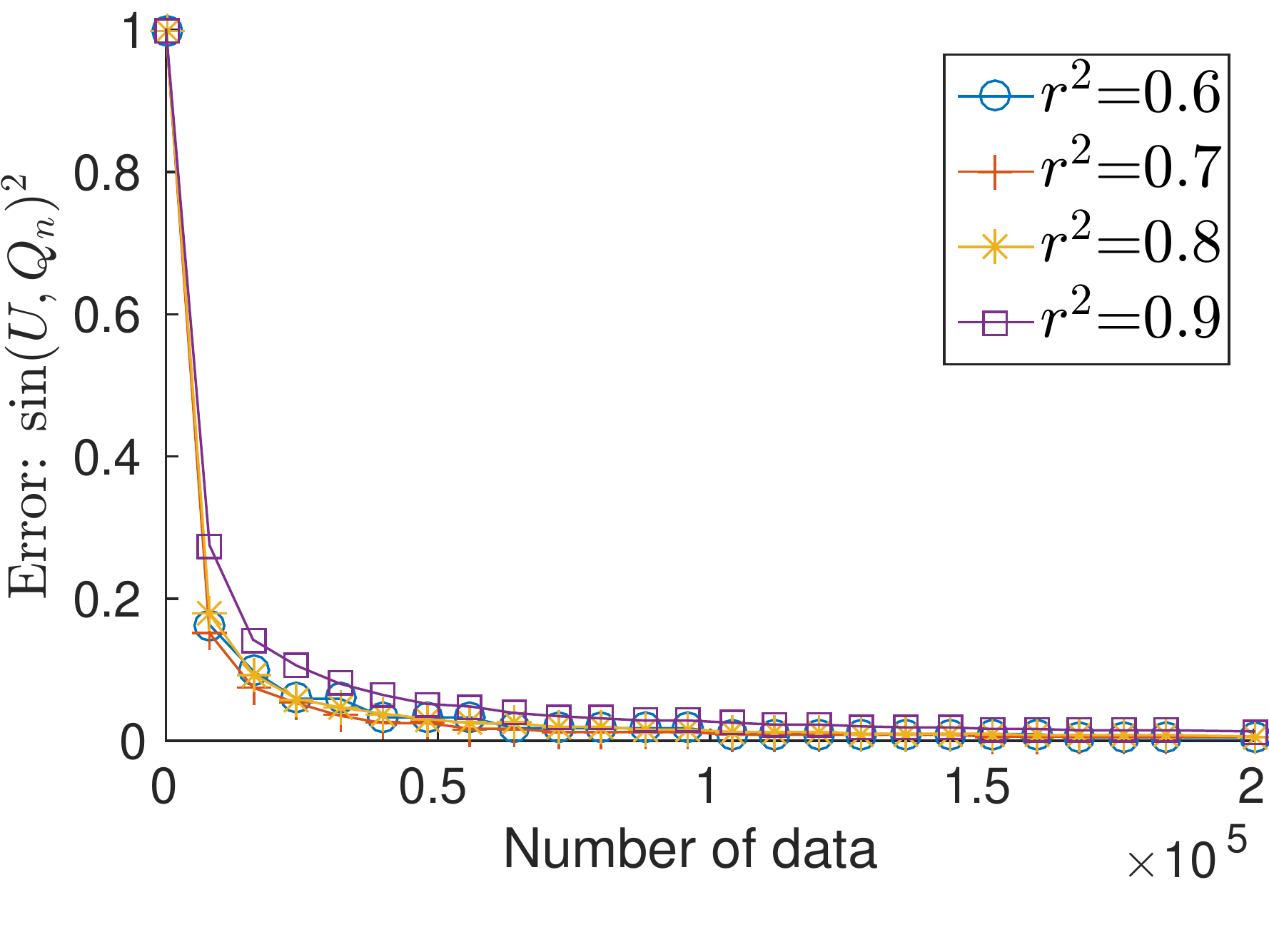} } 
  \caption{Performance of different algorithms on PubMed when $k=4$}
  \label{fig:pm4}
\end{figure}

\begin{figure}[t]
  \centering
  \subfigure[Alecton, $k=10$]{\label{fig:pm_a_10}\includegraphics[width=0.23\textwidth]{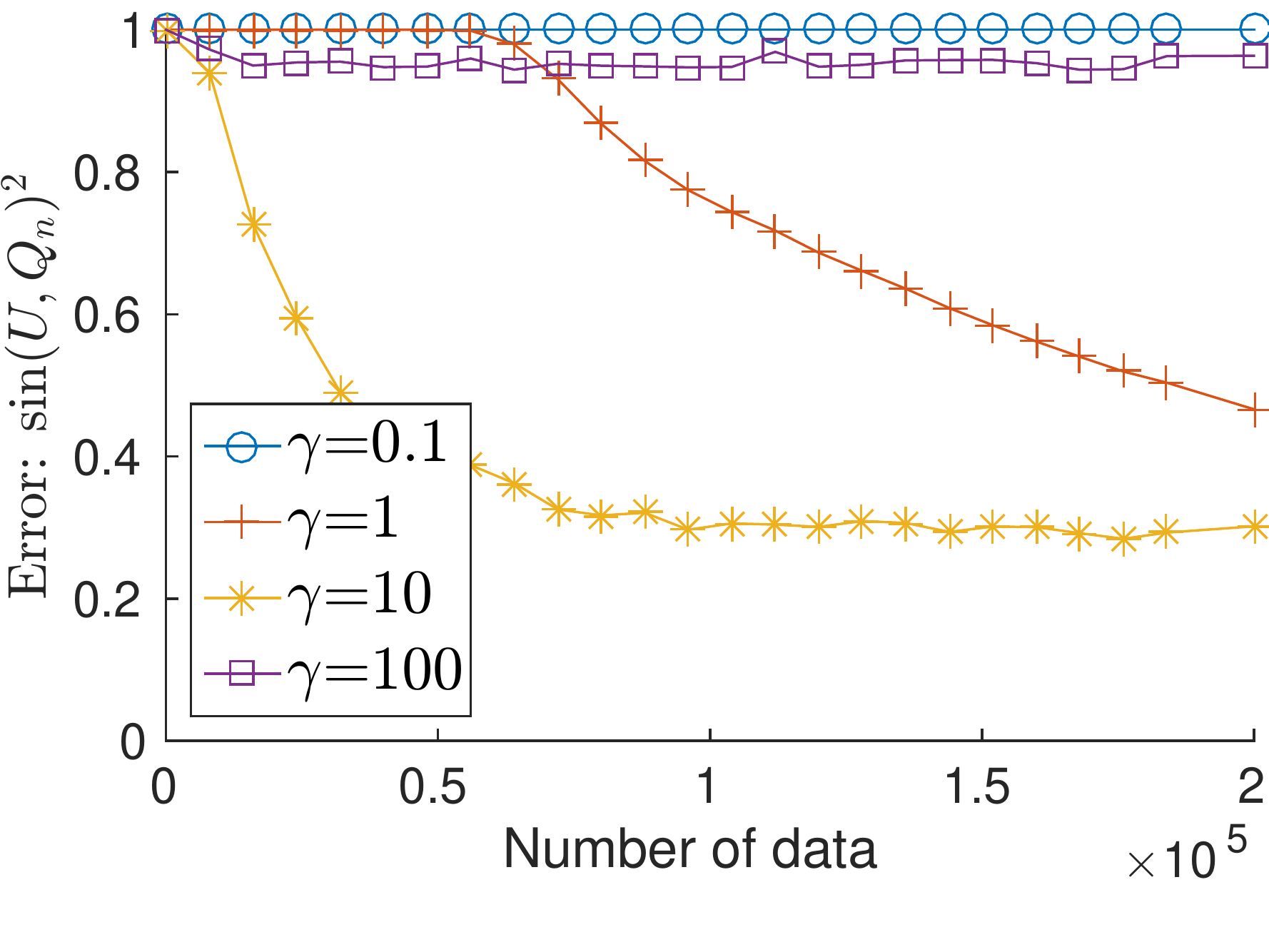} } 
  \subfigure[SPCA, $k=10$]{\label{fig:pm_s_10}\includegraphics[width=0.23\textwidth]{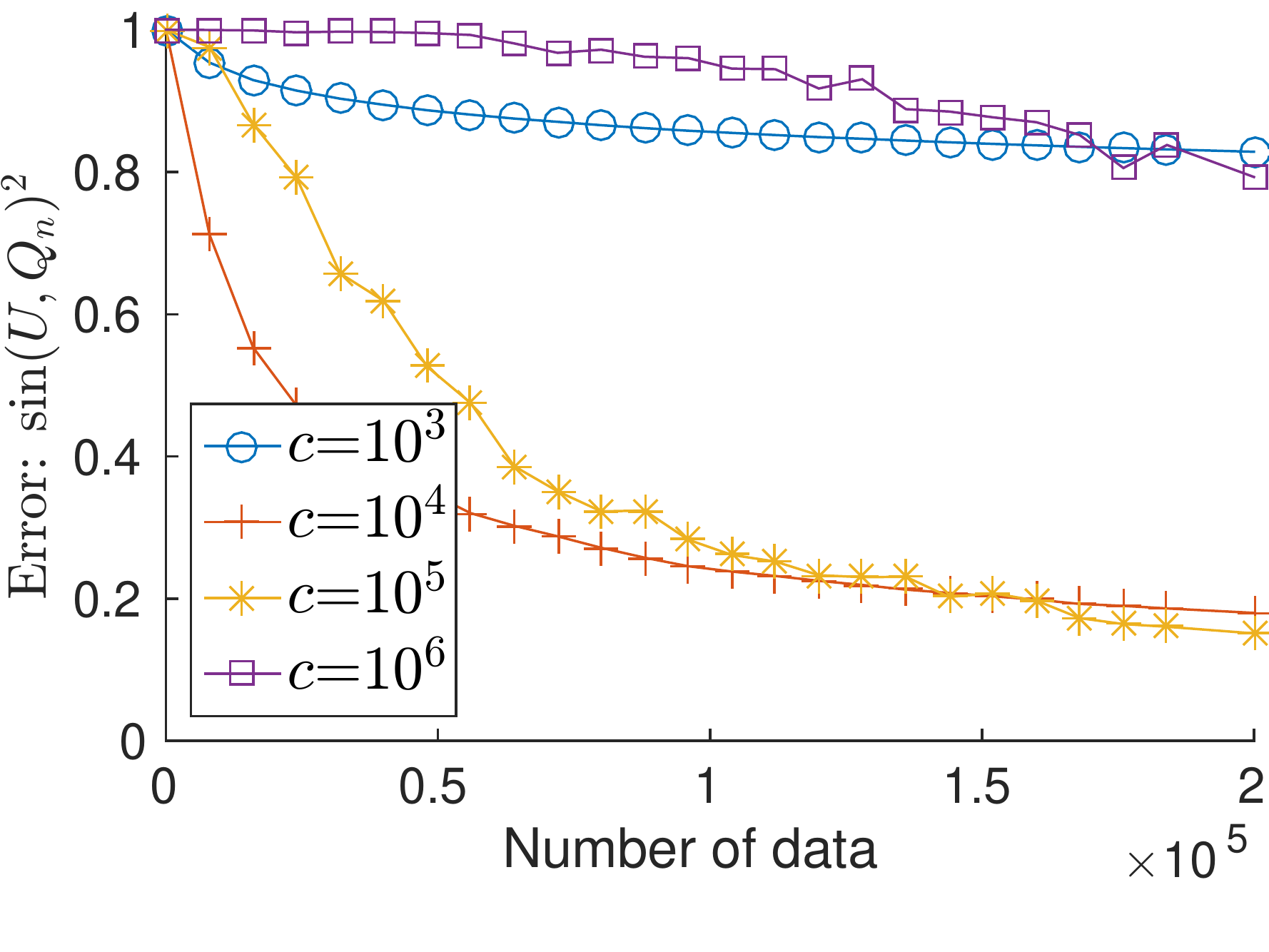} } 
  \subfigure[BPCA, $k=10$]{\label{fig:pm_b_10}\includegraphics[width=0.23\textwidth]{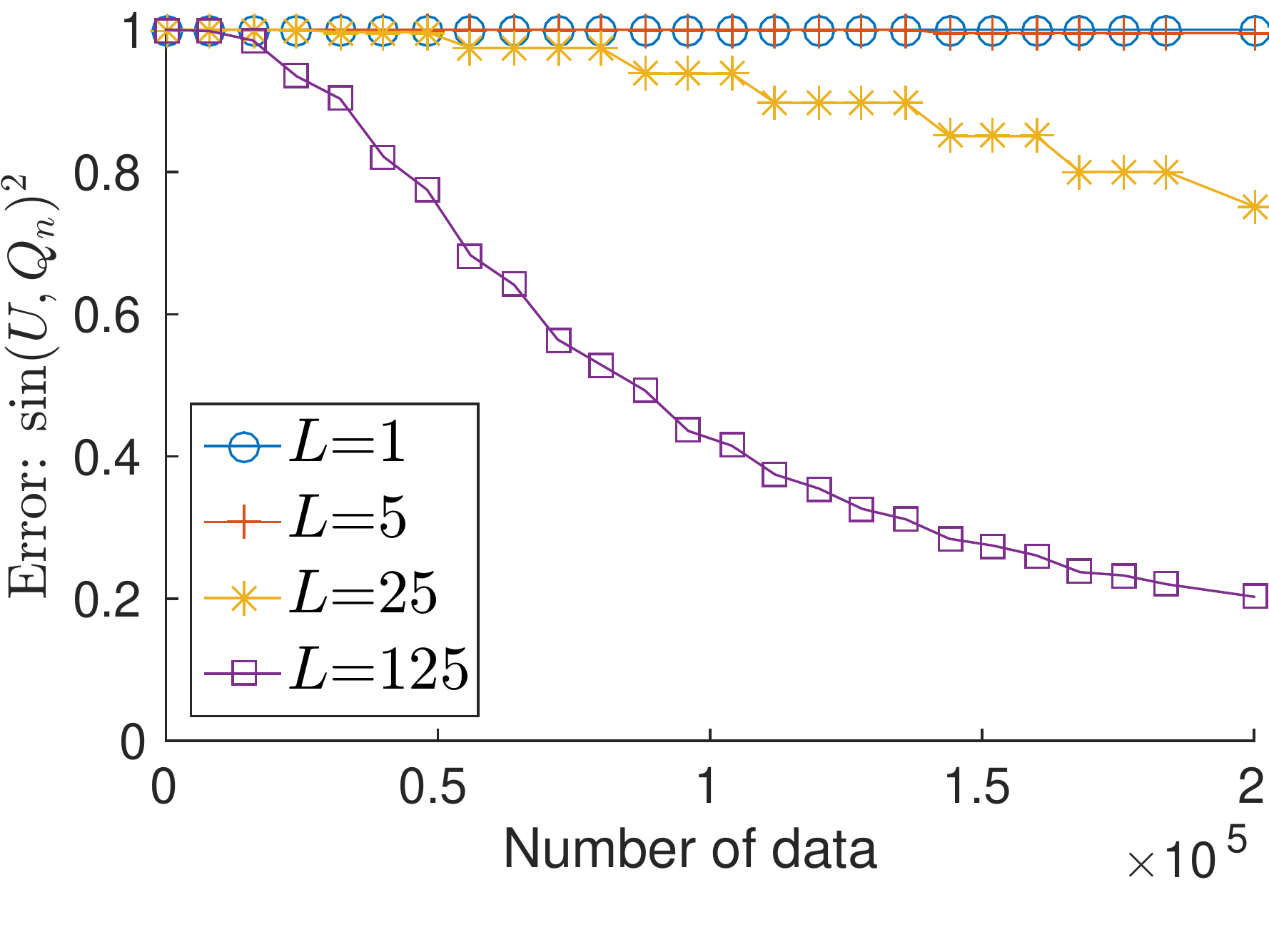} } 
  \subfigure[DBPCA, $k=10$]{\label{fig:pm_d_10}\includegraphics[width=0.23\textwidth]{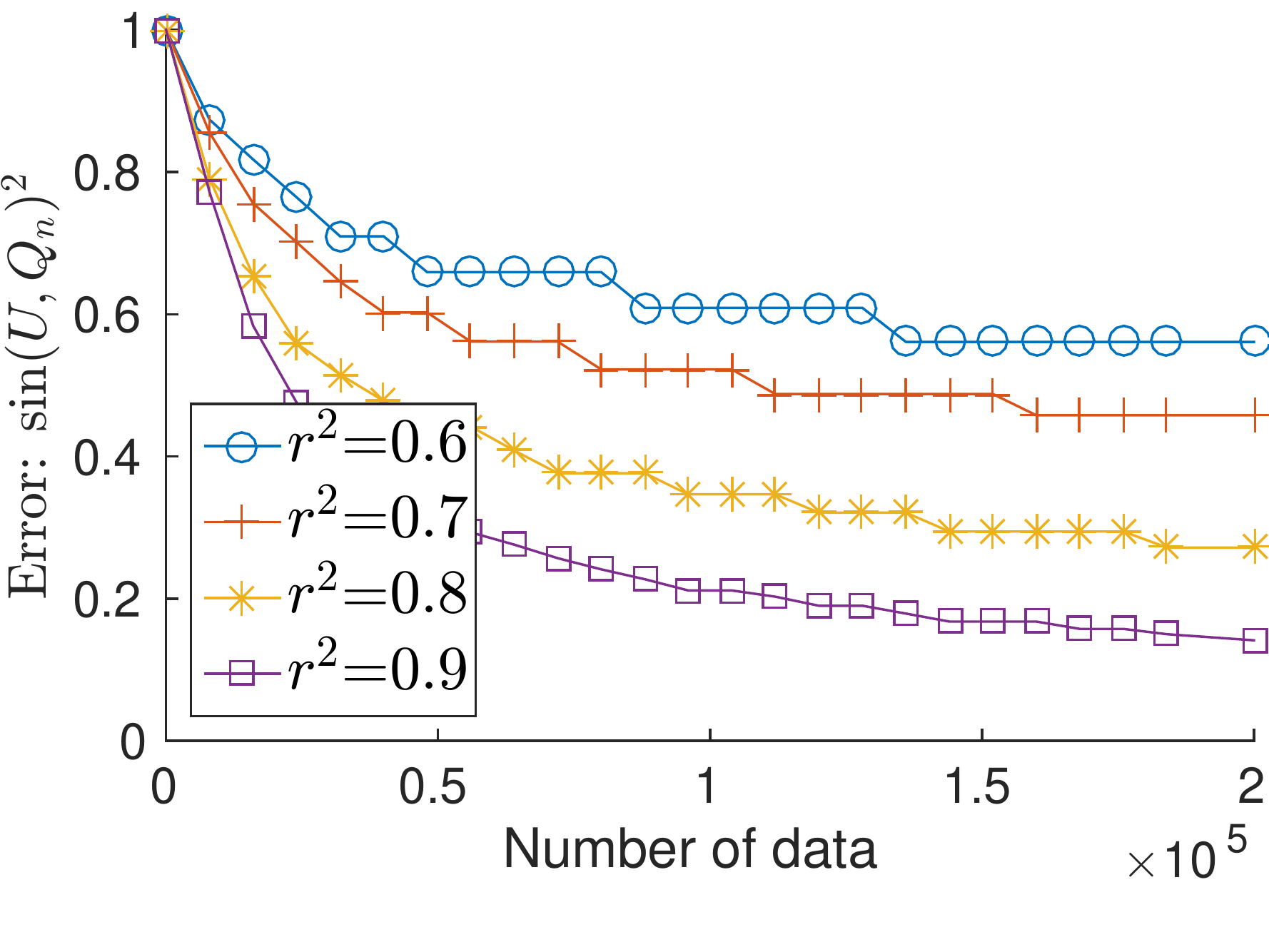} } 
  \caption{Performance of different algorithms on PubMed when $k=10$}
  \label{fig:pm10}
\end{figure}

We run each algorithm $60$ times by randomly generating data streams from the dataset. We consider $\sin(U, Q_n)^2$,
which is the error function used for the convergence analysis, as the performance evaluation criterion. The average
performance on the two datasets for $k = 4$ and $k = 10$ are shown in Figure~\ref{fig:ny4}, Figure~\ref{fig:ny10},
Figure~\ref{fig:pm4} and Figure~\ref{fig:pm10},
respectively. Our experiments on other~$k$ values lead to similar observations and are not included here because of the
space limit.  Also, we report the mean and the standard error of each algorithm with the best parameters in
Tables~\ref{tb:ny4},~\ref{tb:ny10},~\ref{tb:pm4} and~\ref{tb:pm10}.  To visualize the results clearly, we
crop the figures up to $n=200,000$, which is sufficient for checking the convergence of most of the parameter choices on
the algorithms.

\begin{table}[t]
    \centering
  \begin{tabular}{c|c|c}
    \hline
    &  $T=10^5$ & $T=2\times 10^5$ \\
    \hline
    Alecton & $0.232\pm 0.028$  & $0.148\pm 0.008$ \\
    SPCA & $0.159\pm 0.008$ & $0.079\pm 0.006$  \\
    BPCA & $0.234\pm 0.021$ & $0.177\pm 0.012$   \\
    DBPCA & $\mathbf{0.138\pm 0.008}$ & $\mathbf{0.064\pm 0.005}$  \\
    \hline
  \end{tabular}
  \caption{Performance of different algorithms with the best parameter on NYTimes when $k=4$} 
  \label{tb:ny4}
\end{table}

\begin{table}[t]
    \centering
  \begin{tabular}{c|c|c}
    \hline
    &  $T=10^5$ & $T=2\times 10^5$ \\
    \hline
    Alecton & $0.385\pm 0.013$  & $0.386\pm 0.012$\\
    SPCA & $\mathbf{0.170\pm 0.023}$ & $\mathbf{0.102\pm 0.018}$   \\
    BPCA & $0.487\pm 0.042$ & $0.317\pm 0.034$  \\
    DBPCA & $0.207\pm 0.028$ & $0.151\pm 0.022$ \\
    \hline
  \end{tabular}
  \caption{Performance of different algorithms with the best parameter on NYTimes when $k=10$} 
  \label{tb:ny10}
\end{table}

\begin{table}[t]
    \centering
  \begin{tabular}{c|c|c}
    \hline
    &  $T=10^5$ & $T=2\times 10^5$ \\
    \hline
    Alecton & $0.051\pm 0.007$  & $0.042\pm 0.000$ \\
    SPCA & $0.033\pm 0.000$ & $0.022\pm 0.000$   \\
    BPCA & $0.045\pm 0.001$ & $0.044\pm 0.000$   \\
    DBPCA & $\mathbf{0.026\pm 0.000}$ & $\mathbf{0.013\pm 0.000}$  \\
    \hline
  \end{tabular}
  \caption{Performance of different algorithms with the best parameter on PubMed when $k=4$} 
  \label{tb:pm4}
\end{table}

\begin{table}[t]
    \centering
  \begin{tabular}{c|c|c}
    \hline
    &  $T=10^5$ & $T=2\times 10^5$ \\
    \hline
    Alecton & $0.291\pm 0.007$  & $0.292\pm 0.009$   \\
    SPCA & $0.274\pm 0.007$ & $0.190\pm 0.040$   \\
    BPCA & $0.415\pm 0.037$ & $0.203\pm 0.030$   \\
    DBPCA & $\mathbf{0.212\pm 0.024}$ & $\mathbf{0.141\pm 0.031}$ \\
    \hline
  \end{tabular}
  \caption{Performance of different algorithms with the best parameter on PubMed when $k=10$} 
  \label{tb:pm10}
\end{table}

\subsection{Comparison between SPCA and Alecton}
\label{sec:alec}
The main difference between SPCA and Alecton is the rule of determining the learning rate. The learning rate of SPCA will decay along with
the number of iterations, which means it could achieve arbitrarily small error when we have more data. On the other
hand, Alecton needs to pre-specify the desired error to determine a fixed learning rate.
To achieve the same error, from Table~\ref{tb:tb_comp_sgd}, SPCA and Alecton have the same asymptotic convergence rate theoretically.
Next, we aim to study their empirical performance. 

\cite{SaRO15} use a conservative rule to determine the learning rate. The upper bound of the learning rate~$\gamma$
suggested in~\cite{SaRO15} is smaller than $10^{-5}$ for both datasets. However, this conservative and fixed learning
rate scheme takes millions of iterations to converge to the competitive performance with SPCA. Similar results can also
be found in~\cite{SaRO15}. 

Although the suggested learning rate should be small, we still study performance of Alecton with larger learning rates,
which are from $10^{-4}$ to $10^4$. We report the results of $\{10^{-1}, 10^0, 10^1, 10^2\}$, which contain the optimal
choices of the used datasets. 
Obviously, SPCA is generally better than Alecton, such as the case in Figure~\ref{fig:ny4}.
From Tables~\ref{tb:ny4},~\ref{tb:ny10},~\ref{tb:pm4} and~\ref{tb:pm10}, SPCA outperforms
Alecton with the best parameters, which demonstrates the advantage of the decayed learning rate used by SPCA. From
all figures, 
although Alecton with a larger learning rate ($\gamma=10$) has a faster convergence behaviour at the beginning, 
it is stuck at a suboptimal point and can not utilize the new incoming data. The smaller learning rate could usually
results in better performance in the end, but it takes more iterations than the number SPCA needs.

\subsection{Comparison between DBPCA and BPCA}
From Figure~\ref{fig:ny4} and Figure~\ref{fig:ny10}, DBPCA outperforms BPCA under most
parameter choices when $k = 4$, and
is competitive to BPCA when $k = 10$. 
The edge of DBPCA over BPCA is even more remarkable in Figure~\ref{fig:pm4} and Figure~\ref{fig:pm10}.
From the result of the best parameters, DBPCA is significantly better than BPCA by $t$-test at $95\%$ confidence.

BPCA has the similar drawback to Alecton.
As can be observed from
Figures~\ref{fig:ny4},~\ref{fig:ny10},~\ref{fig:pm4} and~\ref{fig:pm10}, 
if $L$ is too small (larger block), BPCA only sees one or two blocks of data within $n = 200,000$, and cannot reduce the error much. 
BPCA typically needs $L > 1$ (smaller blocks) to achieve lower error in the end. 
$L=125$ gives the best performance of BPCA in Figures~\ref{fig:pm4} and~\ref{fig:pm10}.
However, sometimes large~$L$ (small blocks) in BPCA allows reducing the error in the beginning, the error cannot converge to a competitive level in the long run. 
For instance, in Figure~\ref{fig:ny_b_10}, 
$L=125$ converges fast but cannot improve much after $n=50,000$; $L=25$ converges slower but keeps going towards the lowest error after $n=200,000$.
Also, using smaller blocks
cannot ensure reducing the error after each update, and hence BPCA with larger $L$
results in less stable curves even after averaging over $60$ runs.
The results shows the difficulty of setting parameters of BPCA by the strategy proposed in~\cite{MCJ13,HP14}.

On the other hand,
DBPCA achieves better results by using a smaller block in the beginning to make improvements and a larger block later to
further reduce the error. Also,
in both datasets and under all parameter choices, DBPCA 
stably reduces the error after each update, which matches
our theoretical analysis that guarantees error reduction with a high probability.
In addition, DBPCA is quite stable with respect to the choice of $\gamma^2$ across the two datasets, making it easier to tune in practice. The properties
make DBPCA favorable over BPCA in the family of block power methods.

\subsection{Comparison between DBPCA and SPCA}
As observed, DBPCA is less sensitive to the parameter $\gamma$ that corresponds to the theoretical suggestion of
$\max({\ort{\lambda}}/{\lambda}, {1}/{4})^{\frac{1}{4}}$. Somehow SPCA is rather
sensitive to the parameter $c$ that corresponds to the theoretical suggestion of $\frac{c_0}{\lambda - \ort{\lambda}}$.
For instance, setting $c = 10^3$ results in strong performance when $k = 4$
in Figure~\ref{fig:ny_s_4}, but the worst performance when $k = 10$ in Figure~\ref{fig:ny_s_10}.  Similar results can be
observed in Figure~\ref{fig:pm_s_4} and Figure~\ref{fig:pm_s_10} when $c=10^3$. Furthermore, the parameter $c$ in SPCA
directly affects the step size of each gradient descent update. Thus, compared with the best parameter choice, larger
$c$ leads to less stable performance curve, while smaller $c$ sometimes results in significantly slower convergence.
The results suggest that SPCA needs a more careful tuning and/or some deeper studies on proper parameter ranges. 

From Tables~\ref{tb:ny4},~\ref{tb:ny10},~\ref{tb:pm4} and~\ref{tb:pm10}, DBPCA significantly outperforms SPCA in 6 out
of 8 cases by $t$-test under $95\%$ confidence.
The result supports the theoretical study that DBPCA has better converges rate
guarantee than SPCA. 

However, the benefit of SPCA is its immediate use of new data point. DBPCA,
as a representative of the block-power-method family, cannot update the solution until the end of the growing block.
Then, the latter points in the larger blocks may be effectively unused for a long period of time.
For instance, in Figure~\ref{fig:ny10}, DBPCA uses larger blocks than the necessary size.
After $N=150,000$, the block size is near to $20,000$, which is less efficient.

\section{Conclusion}

We strengthen two families of streaming PCA algorithms, and compare the two strengthened families fairly from
both theoretical and empirical sides. For the SGD family, we analyze the convergence rate of the famous SPCA
algorithm for the multiple-principal-component cases without specifying the error in advance; for the family of
block power methods, we propose a dynamic-block algorithm DBPCA that enjoys faster convergence rate than the
original BPCA.  Then, the empirical studies demonstrate that DBPCA not only outperforms BPCA often by
dynamically enlarging the block sizes, but also converges to competitive results more stably than SPCA in many
cases.  Both the theoretical and empirical studies thus justify that DBPCA is the best among the two families,
with the caveat of stalling the use of data points in larger blocks.

Our work opens some new research directions. Empirical results seem to suggest SPCA is competitive to or slightly worse
than DBPCA. It is worth studying whether it is resulted from the substantial difference between $\log 1/\epsilon$ and
$\log \log 1/\epsilon$ or caused by the hidden constants in the bounds.
So one conjecture is that the bound in
Theorem~\ref{thm:itr_conv} can be further improved.  On the other hand, although (\ref{eq:gamma}) suggests $\gamma^2\ge
0.5$, the empirical results show that larger~$\gamma^2$ generally results in better performance. Hence, it is also worth
studying whether the lower bound could be further improved.

\bibliography{paper}
\bibliographystyle{apalike}

\appendix

\section{Proof of Lemma~\ref{lem:rec}} \label{app:rec}

Using the notation $\hat{\bv} = Y_{n-1}\bv/ \|Y_{n-1} \bv\|$ and $A_n = \bx_n \bx_n^\top$, one can follow the analysis
in \cite{fastpca} to show that
$\Phi_n^{(\bv)} \leq \Phi_{n-1}^{(\bv)} + \beta_n - Z_n,$ with
\begin{itemize}
  \item $\beta_n=5\gamma_n^2 + 2\gamma_n^3$,
  \item $Z_n = 2\gamma_n (\hat{\bv}^\top U U^\top A_n \hat{\bv} -  \|U^\top \hat{\bv}\|^2 \hat{\bv}^\top A_n \hat{\bv})$, and
  \item $\E{Z_n|\mathcal{F}_{n-1}}\geq 2\gamma_n(\lambda-\ort{\lambda}) \Phi_{n-1}^{(\bv)} (1-\Phi_{n-1}^{(\bv)})\geq 0$.
\end{itemize}
We omit the proof here as the adaptation is straightforward. It remains to show our better bound on $|Z_n|$. For this, note that
$$|Z_n| \le 2 \gamma_n \left\|\hat{\bv}^\top U U^\top - \|U^\top\hat{\bv}\|^2 \hat{\bv}^\top\right\| \cdot \|A_n \hat{\bv}\|,$$
where $\|A_n \hat{\bv}\| \le 1$ and
\begin{eqnarray*}
\lefteqn{\left\|\hat{\bv}^\top U U^\top - \|U^\top\hat{\bv}\|^2 \hat{\bv}^\top\right\|^2}\\ &=& \|U^\top\hat{\bv}\|^2 - 2 \|U^\top\hat{\bv}\|^4 + \|U^\top\hat{\bv}\|^4\\
&=& \|U^\top\hat{\bv}\|^2 \left(1-\|U^\top\hat{\bv}\|^2\right).
\end{eqnarray*}
As $\|U^\top\hat{\bv}\|^2 \le 1$ and $\left(1-\|U^\top\hat{\bv}\|^2\right) = \Phi_{n-1}^{(\bv)}$, we have
$$|Z_n| \le 2 \gamma_n \sqrt{\Phi_{n-1}^{(\bv)}}.$$

\section{Proof of Lemma~\ref{lem:g1}} \label{app:g1}

Assume that the event $\Gamma_0$ holds and consider any $n \in [n_0, n_1)$. We need the following, which we prove in Appendix~\ref{app:cos}.
\begin{proposition} \label{pro:cos}
For any $n > m$ and any $\bv \in \R^k$,
$$\frac{\|U^\top Y_n \bv\|}{\|Y_n\|} \ge \left(\frac{m}{n}\right)^{3c} \cdot \frac{\|U^\top Y_m \bv\|}{\|Y_m\|}.$$
\end{proposition}
From Proposition~\ref{pro:cos}, we know that for any $\bv \in \Sk$,
$$\frac{\|U^\top Y_n \bv\|}{\|Y_n \bv\|} \ge \frac{\|U^\top Y_n \bv\|}{\|Y_n\|} \ge \left(\frac{n_0}{n}\right)^{3c}
\frac{\|U^\top Y_0 \bv\|}{\|Y_0\|},$$
where $(n_0/n)^{3c} \ge (n_0/n_1)^{3c} \ge (1/c_1)^{3c}$ for the constant $c_1$ given in Remark~\ref{rm:c_1}. As
$Y_0=Q_0$ and $\|Q_0\| = 1 = \|Q_0 \bv\|$, we obtain
$$\frac{\|U^\top Y_n \bv\|}{\|Y_n \bv\|} \ge \frac{\|U^\top Q_0 \bv\|}{c_1^{3c} \|Q_0 \bv\|} \ge \frac{\sqrt{1-\rho_0}}{c_1^{3c}} = \sqrt{\frac{\bar{c}}{c_1^{6c}kd}}.$$
Therefore, assuming $\Gamma_0$, we always have
$$\Phi_n = \max_\bv \left(1- \frac{\|U^\top Y_n \bv\|^2}{\|Y_n \bv\|^2}\right) \le 1 - \frac{\bar{c}}{c_1^{6c}kd} = \rho_1.$$

\subsection{Proof of Proposition~\ref{pro:cos}} \label{app:cos}

Recall that for any $n$, $Y_n = Y_{n-1} + \gamma_n \bx_n \bx_n^\top Y_{n-1}$ and $\|\bx_n \bx_n^\top\| \le 1$. Then for
any $\bv \in \R^k$,
$$\frac{\|U^\top Y_n \bv\|}{\|Y_n\|} \ge \frac{\|U^\top Y_{n-1} \bv\| - \gamma_n \|U^\top Y_{n-1} \bv\|}{\|Y_{n-1}\| + \gamma_n \|Y_{n-1}\|},$$
which is
$$\frac{1 - \gamma_n}{1+\gamma_n} \cdot \frac{\|U^\top Y_{n-1} \bv\|}{\|Y_{n-1}\|} \ge e^{-3\gamma_n} \frac{\|U^\top Y_{n-1} \bv\|}{\|Y_{n-1}\|},$$
using the fact that $1-x \ge e^{-2x}$ for $x \le 1/2$ and $\gamma_n \le 1/2$. Then by induction, we have
$$\frac{\|U^\top Y_n \bv\|}{\|Y_n\|} \ge e^{-3\sum_{t>m}^n \gamma_i} \cdot \frac{\|U^\top Y_m \bv\|}{\|Y_m\|}.$$
The Proposition follows as
$$e^{-3\sum_{t>m}^n \gamma_i} = e^{-3c \sum_{t>m}^n \frac{1}{t}} \ge \left(\frac{m}{n}\right)^{3c}$$
using the fact that $\sum_{t>m}^n \frac{1}{t} \le \int_m^n \frac{1}{x} dx = \ln(\frac{n}{m})$.

\section{Proof of Lemma~\ref{lem:ind}} \label{sec:gamma}

According to Lemma~\ref{lem:rec}, our $\Phi_n^{(\bv)}$'s satisfy the same recurrence relation as the functions $\Psi_n$'s of~\cite{fastpca}. We can therefore have the following, which we prove in Appendix~\ref{app:Phi}.

\begin{lemma} \label{lem:Phi}
Let $\hat{\rho}_i = \rho_i / \lceil e^{5/c_0}\rceil^{c_0 (1-\rho_i)}$. Then for any $\bu \in \Sk$ and $\alpha_i \ge 12c^2/n_{i-1}$,
$$\pr{\sup_{n \ge n_i} \Phi_n^{(\bu)} \ge \hat{\rho}_i + \alpha_i \mid \Gamma_i} \le e^{-\Omega((\alpha_i^2/(c^2\rho_i)) n_{i-1})}.$$
\end{lemma}

Our goal is to bound $\pr{\neg \Gamma_{i+1} | \Gamma_i}$, which is
$$\pr{\exists \bv \in \Sk: \sup_{n_i \le n < n_{i+1}} \Phi_n^{(\bv)} \ge \rho_{i+1} | \Gamma_i}.$$
As discussed before, we cannot directly apply a union bound on the bound in Lemma~\ref{lem:Phi} as there are infinitely
many $\bv$'s in $\Sk$. Instead, we look for a small ``$\epsilon$-net" $\Di_i$ of $\Sk$, with the property that any $\bv \in \Sk$ has some $\bu \in \Di_i$ with $\|\bv-\bu\|\le \epsilon$. Such a $\Di_i$ with $|\Di_i| \le (1/\epsilon)^{\cO(k)}$
is known to exist (see e.g. \cite{Mil86}). Then what we need is that when $\bv$ and $\bu$ are close, $\Phi_n^{(\bv)}$ and $\Phi_n^{(\bu)}$ are close as well. This is guaranteed by the following, which we prove in Appendix~\ref{app:net}.

\begin{lemma} \label{lem:net}
Suppose $\Gamma_i$ happens. Then for any $n \in [n_i,n_{i+1})$, any $\epsilon \le \sqrt{1-\rho_i}/(2c_1^{6c})$, and any $\bu,\bv \in \Sk$ with $\|\bu-\bv\| \le \epsilon$, we have
$$\left|\Phi_n^{(\bv)} - \Phi_n^{(\bu)}\right| \le 16 c_1^{6c} \epsilon / \sqrt{1-\rho_i}.$$
\end{lemma}

According to this, we can choose $\alpha_i=(\rho_{i+1}-\hat{\rho}_i)/2$ and $\epsilon= \alpha_i \sqrt{1-\rho_i} / (16
c_1^{6c})$ so that with $\|\bu-\bv\|\le \epsilon$, we have $|\Phi_n^{(\bv)} - \Phi_n^{(\bu)}| \le \alpha_i$. This means
that given any $\bv\in \Sk$ with $\Phi_n^{(\bv)} \ge \rho_{i+1}$, there exists some $\bu \in \Di_i$ with $\Phi_n^{(\bu)} \ge \rho_{i+1} - \alpha_i = \hat{\rho}_i + \alpha_i$. As a result, we can now apply
a union bound over $\Di_i$ and have
\begin{equation}\label{eq:exp}
\pr{\neg \Gamma_{i+1} | \Gamma_i} \le \sum_{\bu \in \Di_i} \pr{\sup_{n \ge n_i} \Phi_n^{(\bu)} \ge \hat{\rho}_i + \alpha_i \mid \Gamma_i}.
\end{equation}
To bound this further, consider the following two cases.

First, for the case of $i < \pi_1$, we have $\rho_i \ge 3/4$ and $\eta_i = 1 -\rho_i \le 1/4$, so that $$\hat{\rho_i} \le \rho_i e^{-5(1-\rho_i)} = (1-\eta_i) e^{-5 \eta_i} \le e^{-6\eta_i} \le 1 - 3 \eta_i.$$
Then $\alpha_i \ge \left((1-2\eta_i) - (1-3\eta_i)\right)/2 = \eta_i/2$, which is at least $12 c^2 / n_{i-1}$, as $\eta_i \ge \eta_1 \ge \bar{c} /(c_1^{6c}kd)$ and $n_{i-1} \ge n_0 = \hat{c}^c k^3 d^2 \log d$ for a large enough constant $\hat{c}$. Therefore, we can apply Lemma~\ref{lem:Phi} and the bound in (\ref{eq:exp}) becomes
$$\left(c_1^c/\eta_i\right)^{\cO(k)} e^{-\Omega((\eta_i^2/c^2) n_{i-1})} \le \frac{\delta_0}{2(i+1)^2}.$$

Next, for the case of $i \ge \pi_1$, we have $\rho_i \le 3/4$ so that
$$\hat{\rho_i} \le \rho_i / \lceil e^{5/c_0}\rceil^{c_0/4} \le \rho_i / \lceil e^{5/c_0}\rceil^3,$$
as $c_0 \ge 12$ by assumption.
Since $\rho_{i+1} \ge \rho_i/\lceil e^{5/c_0}\rceil^2$,
this gives us $\alpha_i \ge \rho_i (\lceil e^{5/c_0}\rceil^{-2} - \lceil e^{5/c_0}\rceil^{-3})/ 2$, which is at least $12 c^2/n_{i-1}$, as $\rho_i$, according to our choice, is about $c_2(c^3 k \log n_{i-1})/(n_{i-1}+1)$ for a large enough constant $c_2$.
Thus, we can apply Lemma~\ref{lem:Phi} and the bound in (\ref{eq:exp}) becomes
\begin{equation} \label{eq:dev}
\left(c_1^c/\rho_i\right)^{\cO(k)} e^{-\Omega((\rho_i/c^2) n_{i-1})} \le \frac{\delta_0}{2(i+1)^2}.
\end{equation}
This completes the proof of Lemma~\ref{lem:ind}.

\subsection{Proof of Lemma~\ref{lem:Phi}} \label{app:Phi}

By Lemma~\ref{lem:rec}, the random variables $\Phi_n^{(\bv)}$'s satisfy the same recurrence relation of~\cite{fastpca}
for their random variables $\Phi_n$'s. Thus, we can follow their analysis\footnote{In particular, their proofs for Lemma
2.9 and Lemma 2.10.}, but use our better bound on $|Z_n|$, and have the following.

First, when given $\Gamma_i$, we have $|Z_n| \le 2\gamma_n \sqrt{\rho_i}$ for $n_{i-1} \le n < n_i$. Then one can easily
modify the analysis in \cite{fastpca} to show that for any $t \ge 0$,
$$\E{e^{t \Phi_{n_i}^{(\bv)}} | \Gamma_i} \le \exp\left(t \hat{\rho}_i + c^2(6t + 2t^2\rho_i )\left(\frac{1}{n_{i-1}}-\frac{1}{n_i}\right)\right),$$
by noting that $(n_i+1)/(n_{i-1}+1) = \lceil e^{5/c_0} \rceil$ and $n\ge n_0=\hat{c}^c k^3 d^2 \log d$ according to our choice of parameters.

Next, following \cite{fastpca} and applying Doob's martingale inequality, we obtain
\begin{eqnarray*}
\lefteqn{\pr{\sup_{n \ge n_i} \Phi_{n}^{(\bv)} \ge \hat{\rho}_i + \alpha_i | \Gamma_i}}\\
&\le& \E{e^{t \Phi_{n_i}^{(\bv)}} | \Gamma_i} \exp\left(- t(\hat{\rho}_i + \alpha_i) + \frac{c^2 }{n_i}(6t+ 2t^2 \rho_i) \right)\\
&\le& \exp\left(-t\alpha_i + \frac{c^2}{ n_{i-1}}(6t+ 2 t^2 \rho_i)\right)\\
&\le& \exp\left(-\frac{t\alpha_i}{2} + \frac{2c^2 t^2 \rho_i}{n_{i-1}}\right),
\end{eqnarray*}
as $\alpha_i \ge \frac{12c^2}{n_{i-1}}$. Finally, by choosing $t=\frac{\alpha_i n_{i-1}}{8 c^2 \rho_i}$, 
we have the lemma.

\subsection{Proof of Lemma~\ref{lem:net}} \label{app:net}

Assume without loss of generality that $\Phi_n^{(\bv)} \le \Phi_n^{(\bu)}$ (otherwise, we switch $\bv$ and $\bu$), so that
$$\left|\Phi_n^{(\bv)} - \Phi_n^{(\bu)}\right| = \frac{\|U^\top Y_n \bv\|^2}{\|Y_n \bv\|^2} - \frac{\|U^\top Y_n \bu\|^2}{\|Y_n \bu\|^2}.$$
As $\|\bv-\bu\| \le \epsilon$, we have
\begin{equation}\label{eq:frac1}
\frac{\|U^\top Y_n \bv\|}{\|Y_n \bv\|} \le \frac{\|U^\top Y_n \bu\| + \epsilon \|U^\top Y_n\|}{\|Y_n \bu\| - \epsilon \|Y_n\|}.
\end{equation}
To relate this to $\frac{\|U^\top Y_n \bu\|^2}{\|Y_n \bu\|^2}$, we would like to express $\|U^\top Y_n\|$ in terms of $\|U^\top Y_n \bu\|$ and $\|Y_n\|$ in terms of $\|Y_n \bu\|$. For this, note that both $\|U^\top Y_n \bu\|/ \|U^\top Y_n\|$ and $\|Y_n \bu\|/\|Y_n\|$ are at least $\|U^\top Y_n \bu\|/\|Y_n\|$, which by Proposition~\ref{pro:cos} is at least
\begin{equation}\label{eq:frac2}
\left(\frac{n_{i-1}}{n}\right)^{3c} \frac{\|U^\top Y_{n_{i-1}} \bu\|}{\|Y_{n_{i-1}}\|} \ge c_1^{-6c} \frac{\|U^\top Y_{n_{i-1}} \bu\|}{\|Y_{n_{i-1}}\|},
\end{equation}
using the fact that $n_{i-1}/n \ge n_{i-1}/n_{i+1} \ge 1/c_1^2$. Then as $Y_{n_{i-1}} = Q_{n_{i-1}}$ and $\|Q_{n_{i-1}}\| = \|Q_{n_{i-1}} \bu\|$, the righthand side of (\ref{eq:frac2}) becomes
$$c_1^{-6c} \frac{\|U^\top Q_{n_{i-1}} \bu\|}{\|Q_{n_{i-1}} \bu\|} = c_1^{-6c} \sqrt{1-\Phi_{n_{i-1}}^{(\bu)}} \ge c_1^{-6c}\sqrt{1-\rho_i},$$
given $\Gamma_i$. What we have obtained so far is a lower bound for both $\|U^\top Y_n \bu\|/ \|U^\top Y_n\|$ and $\|Y_n \bu\|/\|Y_n\|$. Plugging this into (\ref{eq:frac1}), with $\hat{\epsilon} = \epsilon c_1^{6c} / \sqrt{1-\rho_i}$, we get
$$\frac{\|U^\top Y_n \bv\|}{\|Y_n \bv\|} \le \frac{\|U^\top Y_n \bu\| (1+ \hat{\epsilon})}{\|Y_n \bu\| (1-\hat{\epsilon})}.$$
As a result, we have
$$\left|\Phi_n^{(\bv)} - \Phi_n^{(\bu)}\right| \le \frac{\|U^\top Y_n \bu\|^2}{\|Y_n \bu\|^2} \left(\frac{(1+\hat{\epsilon})^2}{(1-\hat{\epsilon})^2} - 1\right) \le 16\hat{\epsilon},$$
since $\frac{(1+\hat{\epsilon})^2}{(1-\hat{\epsilon})^2} - 1 \le \frac{4\hat{\epsilon}}{(1-\hat{\epsilon})^2} \le 16\hat{\epsilon}$ for $\hat{\epsilon} \le 1/2$.

\section{Proof of Lemma~\ref{lem:err}} \label{app:err}
As $\cos(U,Q_{i-1})^2 = \frac{1}{1+ \tan(U,Q_{i-1})^2} \ge \frac{1}{1+ \varepsilon_{i-1}^2} \ge \beta_i^2$,
we have $\|G_i\| \le \triangle \beta_i \le \triangle \cos(U,Q_{i-1}).$
Thus, we can apply Lemma~\ref{lem:next} and have
$$\tan (U,A Q_{i-1}+G_i) \le \max(\beta_i, \max(\beta_i, \gamma) \varepsilon_{i-1}),$$
which is at most $\max(\beta_i, \gamma \varepsilon_{i-1}) \le \gamma \varepsilon_{i-1} = \varepsilon_i$. The lemma follows as $\tan (U,Q_i) = \tan (U,A Q_{i-1}+G_i)$.

\section{Proof of Lemma~\ref{lem:dev}}\label{app:dev}
Let $\rho = \triangle\beta_i$ and note that $\|G_i\| \le \|A - F_i\|$, where $F_i$ is the average of $|I_i|$ i.i.d.~random matrices, each with mean $A$. Recall that $\|A\|\le 1$ by Assumption~\ref{as:A}. Then from a matrix Chernoff bound, we have
$$\pr{\|G_i\| > \rho} \le \pr{\|A - F_i\| > \rho} \le d e^{-\Omega(\rho^2 |I_i|)} \le \delta_i,$$
for $|I_i|$ given in (\ref{eq:n_i}).

\section{Proof of Lemma~\ref{lem:n_sample}}\label{app:n_sample}

Let $L$ be the iteration number such that $\varepsilon_{L-1} > \varepsilon$ and $\varepsilon_L \le \varepsilon$.
Note that with $\varepsilon_L = \varepsilon_0 \gamma^L = \varepsilon_0 (1-(\lambda-\td{\lambda})/\lambda)^{L/4} \le \varepsilon_0 e^{-L(\lambda-\td{\lambda})/(4\lambda)}$, we can have
$$L \le \cO\left(\frac{\lambda}{\lambda-\td{\lambda}}\log\frac{\varepsilon_0}{\varepsilon}\right)
\le \cO\left(\frac{\lambda}{\lambda-\td{\lambda}}\log\frac{d}{\varepsilon}\right).$$

As the number of samples in iteration $i$ is
$$|I_i| = \cO\left(\frac{\log(d/\delta_i)}{(\lambda-\td{\lambda})^2 \beta_i^2}\right) \le \cO\left(\frac{\log (di)}{(\lambda-\td{\lambda})^2 \beta_i^2}\right),$$
the total number of samples needed is
$$\sum_{i=1}^L |I_i| \le \cO\left(\frac{\log (dL)}{(\lambda-\td{\lambda})^2}\right) \cdot \sum_{i=1}^L \frac{1}{\beta_i^2}.$$
With $\beta_i = \min(\gamma/\sqrt{1+\varepsilon_{i-1}^2}, \gamma\varepsilon_{i-1})$, one sees that for some
$i_0 \le \cO(\log d)$, $\beta_i = \gamma/\sqrt{1+\varepsilon_{i-1}^2}$ when $i \le i_0$ and $\beta_i = \gamma\varepsilon_{i-1} = \varepsilon_i$ when $i > i_0$. This implies that
\begin{equation}\label{eq:beta}
\sum_{i=1}^L \frac{1}{\beta_i^2} = \sum_{i=1}^{i_0} \frac{1+\varepsilon_{i-1}^2}{\gamma^2} + \sum_{i=i_0+1}^L \frac{1}{\varepsilon_i^2},
\end{equation}
where the first sum in the righthand side of (\ref{eq:beta}) is
$$\frac{i_0}{\gamma^2} + \sum_{i=1}^{i_0} \varepsilon_0^2 \gamma^{2i-4} \le \frac{\cO(\log d)}{\gamma^2} + \frac{\varepsilon_0^2}{\gamma^2(1-\gamma^2)},$$
while the second sum is
$$\sum_{i=i_0+1}^L \frac{\gamma^{2(L-i)}}{\varepsilon_L^2} \le \frac{1}{(1-\gamma^2)\varepsilon_L^2} \le \frac{1}{\gamma^2(1-\gamma^2) \varepsilon^2}$$
using the fact that $\varepsilon_L = \gamma \varepsilon_{L-1} \ge \gamma \varepsilon$. Since
$\gamma^2 = \left(1-\frac{\lambda-\td{\lambda}}{\lambda}\right)^{1/2} \le 1-\frac{\lambda-\td{\lambda}}{2\lambda},$
we have $\frac{1}{1-\gamma^2} \le \frac{2\lambda}{\lambda-\td{\lambda}}$, and since $\lambda \le \cO(\td{\lambda})$, we also have $\frac{1}{\gamma^2} \le \cO(1)$. Moreover, as we assume that $\varepsilon \le 1/\sqrt{kd}$,
we can conclude that the total number of samples needed is at most
$$\sum_{i=1}^L |I_i| \le \cO\left(\frac{\log (dL)}{(\lambda-\td{\lambda})^2}\right) \cdot \cO\left(\frac{\lambda}{(\lambda-\td{\lambda})\varepsilon^2}\right) \le \cO\left(\frac{\lambda\log (dL)}{\varepsilon^2(\lambda-\td{\lambda})^3}\right).$$

\end{document}